\documentclass[lettersize,journal]{IEEEtran}
\usepackage{amsmath,amsfonts}
\usepackage{algorithmic}
\usepackage{array}
\usepackage{textcomp}
\usepackage{url}
\usepackage{verbatim}
\usepackage{graphicx}
\def\BibTeX{{\rm B\kern-.05em{\sc i\kern-.025em b}\kern-.08em
    T\kern-.1667em\lower.7ex\hbox{E}\kern-.125emX}}
\usepackage{balance}
\usepackage{multirow}
\usepackage{algorithm}
\usepackage{graphicx}
\usepackage{cite}
\usepackage{stfloats}
\usepackage{subfigure}
\usepackage{hyperref}
\begin{document}
\title{CoordLight: Learning Decentralized Coordination for Network-Wide Traffic Signal Control}
\author{Yifeng Zhang, Harsh Goel, Peizhuo Li, Mehul Damani, Sandeep Chinchali, Guillaume Sartoretti
\thanks{This work is supported by A*STAR, CISCO Systems (USA) Pte. Ltd and National University of Singapore under its Cisco-NUS Accelerated Digital Economy Corporate Laboratory (Award I21001E0002).}
\thanks{Yifeng Zhang, Peizhuo Li, and Guillaume Sartoretti are with the Department of Mechanical Engineering, National University of Singapore (E-mail: yifeng@u.nus.edu, e0376963@u.nus.edu, guillaume.sartoretti@nus.edu.sg)}
\thanks{Harsh Goel and Sandeep Chinchali are with the Chandra Department of Electrical and Computer Engineering, The University of Texas at Austin (E-mail: harshg99@utexas.edu, sandeepc@utexas.edu)}
\thanks{Mehul Damani is with the Department of Electrical Engineering and Computer Science, Massachusetts Institute of Technology (E-mail:damanimehul24@gmail.com)}}

\maketitle

\begin{abstract}
\label{abstract}
Adaptive traffic signal control (ATSC) is crucial in alleviating congestion, maximizing throughput and promoting sustainable mobility in ever-expanding cities.
Multi-Agent Reinforcement Learning (MARL) has recently shown significant potential in addressing complex traffic dynamics, but the intricacies of partial observability and coordination in decentralized environments still remain key challenges in formulating scalable and efficient control strategies.
To address these challenges, we present CoordLight, a MARL-based framework designed to improve intra-neighborhood traffic by enhancing decision-making at individual junctions (agents), as well as coordination with neighboring agents, thereby scaling up to network-level traffic optimization.
Specifically, we introduce the Queue Dynamic State Encoding (QDSE), a novel state representation based on vehicle queuing models, which strengthens the agents' capability to analyze, predict, and respond to local traffic dynamics.
We further propose an advanced MARL algorithm, named Neighbor-aware Policy Optimization (NAPO). It integrates an attention mechanism that discerns the state and action dependencies among adjacent agents, aiming to facilitate more coordinated decision-making, and to improve policy learning updates through robust advantage calculation.
This enables agents to identify and prioritize crucial interactions with influential neighbors, thus enhancing the targeted coordination and collaboration among agents. 
Through comprehensive evaluations against state-of-the-art traffic signal control methods over three real-world traffic datasets composed of up to 196 intersections, we empirically show that CoordLight consistently exhibits superior performance across diverse traffic networks with varying traffic flows.
The code is available at \href{https://github.com/marmotlab/CoordLight}{https://github.com/marmotlab/CoordLight}.

\end{abstract}


\section{Introduction}
\label{introduction}
Rapid urbanization has resulted in increasing travel demands in ever-expanding cities, with the resulting traffic congestion posing a significant challenge, mainly in terms of extended travel times, increased environmental pollution, and reduced quality of life.
As a result, adaptive traffic signal control (ATSC)~\cite{lowrie1990scats,little1981maxband, varaiya2013max} has emerged as an effective approach to help optimize urban traffic flows by dynamically updating signal traffic phases and/or timings in response to real-time traffic conditions, offering significant advantages over traditional fixed-time methods.
In particular, recent multi-agent reinforcement learning (MARL) techniques have shown promising outcomes for ATSC~\cite{li2016traffic, garg2018deep, liang2019deep, chu2019multi}.
Centralized RL methods~\cite{prashanth2011reinforcement} were first proposed, in which a single decision-making entity learns to optimize global objectives for all intersections at once.
However, faced with growing urban networks, these methods struggle with the exponential growth of the joint state-action space and the ensuing delay from data centralization.
More recently, the community has turned to decentralized MARL approaches~\cite{chu2019multi, wei2019colight, wei2019presslight, chen2020toward}. These approaches adopt an independent learning framework where each individual traffic intersection is treated as a learning agent, relying on real-time, local traffic data to make control decisions (often, phase selection for a fixed duration).

However, decentralized control also comes with critical challenges. 
One critical issue lies in the lack of a state definition that can efficiently capture the dynamic behaviors of traffic situations. 
In MARL, most of the widely used state definitions are based on lane-specific feature vectors. Numerous definitions have proven effective in describing the traffic conditions at intersections, such as vehicle count~\cite{wei2019colight, attentionlight, goel2023sociallight}, pressure~\cite{wei2019presslight, chen2020toward}, or other enhanced variants such as Efficient Pressure (EP)~\cite{wu2021efficient} and Advanced Traffic State (ATS)~\cite{zhang2022expression}.
Nonetheless, these state definitions fail to provide a full picture of the traffic dynamics, making it difficult for RL agents to learn and adapt robust control strategies in ever-changing traffic conditions.
Additionally, in decentralized settings, agents can only make decisions based on limited traffic information due to partial observability of the global traffic state, which leads to a surge of myopic and self-interested behaviors that inhibit efficient network-wide cooperation. 
In response, recent research has focused on improving agent cooperation and coordination via either advanced feature extraction, communication mechanisms~\cite{wei2019colight, wang2020stmarl, ma2c, lin2023denselight} or the implementation of well-designed reward structures~\cite{wei2019presslight, chen2020toward, lin2023denselight, ma2c}.
Nevertheless, the intricacies of agent interactions and dependencies further exacerbate the issue of instability and scalability within the multi-agent environment. Such dynamic interplays typically lead to an unstable training process and ultimately yield sub-optimal solutions.

\begin{figure*}[t]
\centering
\includegraphics[width=0.9\textwidth,height=0.5\textwidth, trim=80 140 160 100, clip]{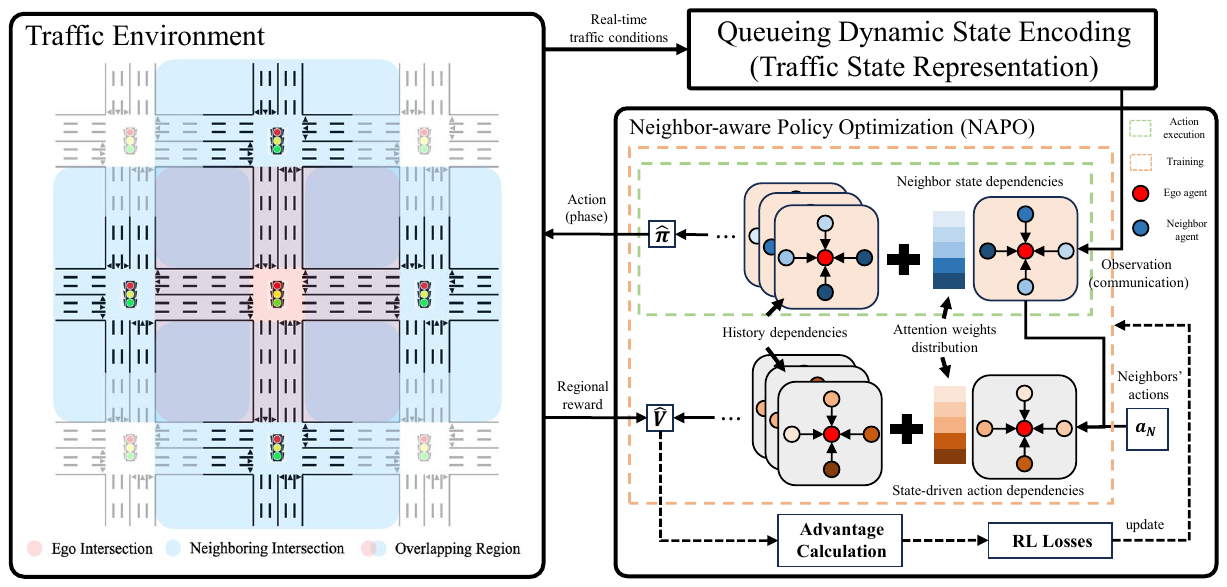}
\vspace{-1.5cm}
\caption{Overall learning framework of CoordLight, which introduces a novel state definition, Queueing Dynamic State Encoding (QDSE) for fine-grained traffic representation, and a Neighbor-aware Policy Optimization (NAPO) algorithm to learn efficient coordination strategies among neighboring agents.}
\vspace{-0.2cm}
\label{framework}
\end{figure*}

To tackle these challenges, we present CoordLight, a novel decentralized MARL framework, as shown in Fig.~\ref{framework}, designed to enhance network-wide traffic signal control.
Our approach introduces a new state definition, called Queue Dynamic State Encoding (QDSE), which integrates the essential lane features from a vehicle queueing dynamics model to capture current traffic conditions, including counts of stopped, entering, and leaving vehicles, and to estimate upcoming congestion based on the position and count of moving vehicles. QDSE significantly enhances the agents' abilities to discern and respond to current and potential congestion by analysing traffic flow dynamics.
We further propose Neighbor-aware Policy Optimization (NAPO), an advanced MARL algorithm designed to facilitate scalable and efficient coordination learning.
Specifically, we employ an attention mechanism to discern the dynamic dependencies among adjacent agents by adaptively identifying influential neighbors based on their state and action information. This effectively decouples the intricate mutual influences within neighborhoods, thereby enabling more targeted coordination and collaboration among agents.
By integrating the neighbors' weighted influence into the value function, we derive a modified advantage estimate that accurately assesses the true contributions of agents' actions towards optimizing their regional objectives, thus enhancing policy learning and leading to a more stable training process.
To realize such neighbor-aware learning, we design an augmented actor-critic network with an attention-based spatio-temporal network as the backbone, aiming to capture vital insights into neighboring states across spatial and temporal dimensions for improved decision-making. 
Additionally, we incorporate a privileged local critic network, designed in an encoder-decoder fashion, aiming to aggregate the state-action dynamics of adjacent agents into the value function.
These enhancements deepen the understanding of potential impacts from neighboring agents, thus promoting the learning of a more adaptive coordinated strategy within our dynamic multi-agent system.

By conducting extensive simulations on a range of benchmark traffic networks composed of up to 196 intersections, using the open-source CityFlow simulator~\cite{tang2019cityflow}, we show that CoordLight consistently outperforms established state-of-the-art learning-based ATSC baselines across diverse traffic networks under the key evaluation metric - average travel time.
From our results, we show that accurately representing complex traffic conditions and dynamically capturing inter-dependencies between states and actions of neighboring agents effectively mitigates the impacts of local observability and environmental instability, thereby leading to improved network-wide performance.
Finally, through a series of ablation studies, we rigorously establish the effectiveness of our proposed two enhancements - QDSE-based state representation, and neighbor-aware policy optimization/actor-critic architecture. 

In summary, the main contributions of this paper are:
\begin{itemize}
\item [(1)] We present a novel state representation, named Queue Dynamic State Encoding (QDSE), which incorporates the vital lane features derived from a queueing dynamics model, to enhance the agents' abilities to analyze, predict and respond to the impending congestion.  
\item [(2)] We further introduce Neighbor-aware Policy Optimization (NAPO), a fully decentralized MARL algorithm, which aims to stabilize training process and enhance coordination among agents by identifying key state-action dependencies with influential neighboring agents.
\item [(3)] We evaluate our proposed learning framework, CoordLight over three real-world traffic datasets composed of up to 196 intersections and demonstrate CoordLight consistently outperforms other state-of-the-art ATSC methods in the metric of average travel time.
\end{itemize}

The remainder of this paper is organized as follows: Section.~\ref{related_works} reviews the literature on traffic signal control. Section.~\ref{background} details traffic terminology, MARL concepts, and the RL formulation. The proposed state representation, Queueing Dynamic State Encoding (QDSE), along with the MARL algorithm, Neighbor-aware Policy Optimization (NAPO) and its network architecture, are presented in Section.~\ref{method:QDSE} and Section.~\ref{method:NAPO} respectively. Simulation experiments and their results are discussed in Section.~\ref{experiments}. Finally, Section.~\ref{conclusion} concludes the paper and outlines directions for future work.

\section{Related Work}
\label{related_works}
Conventional traffic signal control methods can be broadly categorized into fixed-time and adaptive based on their ability to adapt/react to real-time traffic conditions. 
Among fixed-time methods, the Webster method~\cite{koonce2008traffic} provides an analytical solution for determining the optimal cycle length at intersections, while GreenWave~\cite{roess2004traffic} focuses on optimizing timing offsets between intersections to reduce vehicle stops. 
In contrast, adaptive methods tailor control to the existing traffic conditions.
The Sydney Coordinated Adaptive Traffic System (SCATS)~\cite{lowrie1990scats} is a widely-deployed (but closed-source), adaptive method that selects from a set of pre-defined signal plans according to a defined performance measure.
A recent advancement, Max-pressure control~\cite{varaiya2013max} aims to balance queue lengths amongst intersections to improve throughput.

Given the successes of RL in controlling single-traffic intersections~\cite{genders2016using,mousavi2017traffic,wei2018intellilight,van2016coordinated,li2016traffic,aslani2017adaptive,casas2017deep, ma2021deep}, researchers have now shifted their focus towards developing methods for coordinating networks of traffic intersections.
Here, learning frameworks can be categorized into two types: centralized and decentralized methods. 
Centralized methods supervise all intersections using one decision-making entity, learning a global policy that maps the agents' joint states to their joint action vector~\cite{prashanth2011reinforcement, van2016coordinated, xie2020iedqn}.
However, as the number of agents increases, the combined state-action space grows exponentially.
This results in substantial challenges due to the limitations in scalability and training complexity, making its application in large urban networks difficult.
In contrast, numerous decentralized methods~\cite{balaji2010urban, aslani2017adaptive, mannion2016experimental, xiong2019learning, brys2014distributed,chu2019multi, zhu2023metavim} propose to assign an independent learning agent to each intersection which makes a decision based on its observed local traffic conditions instead of the global traffic state, considering other agents as part of the environment dynamics.
However, these decentralized learning methods often encounter challenges in adapting to complex, dynamic environments due to the lack of effective coordination and to their inherently reactive nature, making it difficult to exhibit the longer-term and larger-scale cooperative behaviors needed for efficient city-wide traffic optimization.

To address the challenges in coordinated ATSC learning, the community has proposed improved reward structures to promote coordination.
PressLight~\cite{wei2019presslight} and MPLight~\cite{chen2020toward} introduced rewards to minimize intersection pressure, improving coordination with neighboring intersections. 
Liu et al.~\cite{liu2021learning} further proposed $\gamma-$Reward and $\gamma-$Attention-Reward that uses spatio-temporal information in the replay buffer to amend the rewards for improved coordination from a hindsight perspective.
Complementing the focus on the reward structure, there has been a significant interest in augmenting the observation space of agents to enhance coordination amongst agents in partially observable settings. For instance, MA2C~\cite{chu2019multi} and NeurComm~\cite{chu2020multi} focused on neighborhood-level cooperation, enriching an ego agent's rewards and observations with data from neighboring agents, facilitated by a spatial discount factor. 
NC-HDQN~\cite{zhang2022neighborhood} suggested optimizing strategies with information weighted by the correlation degrees between agents, calculated using empirical rules or Pearson correlation coefficients, to enhance cooperation among agents.
Furthermore, Zhang et al.~\cite{zhang2022expression} introduced "Advanced-MP," a technique that incorporates both moving and queuing vehicles for signal phase decisions. 
Additionally, CoLight~\cite{wei2019colight} utilized Graph Attention Networks (GATs) in traffic signal control to enhance agent communication and cooperation, followed by STMARL~\cite{wang2020stmarl} which improved decision-making by extracting spatio-temporal features through a combination of Graph Neural Networks (GNNs) and Recurrent Neural Networks (RNNs). 
Lin et al.~\cite{lin2023temporal} proposed TeDA-GCRL, which utilizes a novel graph convolutional architecture that treats each lane as a node to capture intersection relations, along with two temporal-aware rewards to enhance the overall traffic performance. In parallel to this development, Zhu et al.~\cite{zhu2023metavim} presented MetaVIM, a meta-learning approach that improves the policy generalizability of agents across different neighborhood sizes by incorporating latent variables, and introduces an intrinsic reward to stabilize the training process in dynamic traffic environments.
While these methods mainly focused on improving coordination through advanced communication or feature extraction mechanisms, they did not fully address the complex behavioral interactions among traffic agents.

Recent works that rely on Centralized Training and Decentralized Execution (CTDE) framework, such as VDN~\cite{van2016coordinated}, COMA~\cite{coma}, QMIX~\cite{rashid2018qmix}, and MAAC~\cite{iqbal2019actor}, have achieved significant advancements in promoting cooperative and collaborative learning within MARL.
For instance, Song et al.~\cite{song2024cooperative} enhanced COMA by incorporating a scheduler module that facilitates efficient information exchange among agents, which improves cooperation in dynamic multi-agent environments.
Zhou et al.~\cite{zhou2024cooperative} proposed the MICDRL framework, which advances the CTDE paradigm for cooperative ATSC by introducing an incentive communication mechanism to enable effective message exchange. Additionally, it improves spatial and temporal traffic dynamics integration by using recurrent neural networks for temporal patterns and mutual information-based modeling for spatial interactions.
These CTDE methods, which often rely on information centralization during training (e.g., centralized critics, mixing networks), contrast with the decentralized methods with local critics that use only local information for learning and decision-making.
However, the reliance of CTDE methods on global information for critic training, even as actions are executed based on local observations, poses notable challenges for developing scalable control strategies in complex, large-scale MARL environments~\cite{gronauer2022multi}. 
In response, Goel et al.~\cite{goel2023sociallight} presented a decentralized cooperation approach, enabling agents to evaluate and optimize their contributions to local traffic performance within neighborhoods to improve global coordination. 
Nonetheless, these approaches, primarily focusing on variance-reducing mechanisms in value estimates for cooperative learning, often tend to overlook the crucial role of neighbor awareness (i.e., the spatial and temporal dependencies among adjacent agents) in the decision-making process.

\section{Background}
\label{background}
\subsection{Traffic Terminology}
\noindent \textbf{\textit{Definition 1 (Incoming and outgoing lanes)}}: 
Incoming lanes at an intersection are where traffic enters, while outgoing lanes are where traffic exits. We denote the set of incoming lanes and outgoing lanes of an intersection as $\mathcal{L}_{in}$ and $\mathcal{L}_{out}$ respectively.

\noindent \textbf{\textit{Definition 2 (Traffic movement)}}:
A traffic movement is a distinct trajectory for vehicles to traverse the intersection, i.e., from an incoming lane to a connected outgoing lane. The traffic movement between lane $l_{in}$ and lane $l_{out}$ is defined as $m(l_{in}, l_{out})$, where $l_{in} \in \mathcal{L}_{in}$ and $l_{out} \in \mathcal{L}_{out}$.

\begin{figure}[tb]
\centering
\includegraphics[width=\linewidth,trim=150 140 120 120,clip]{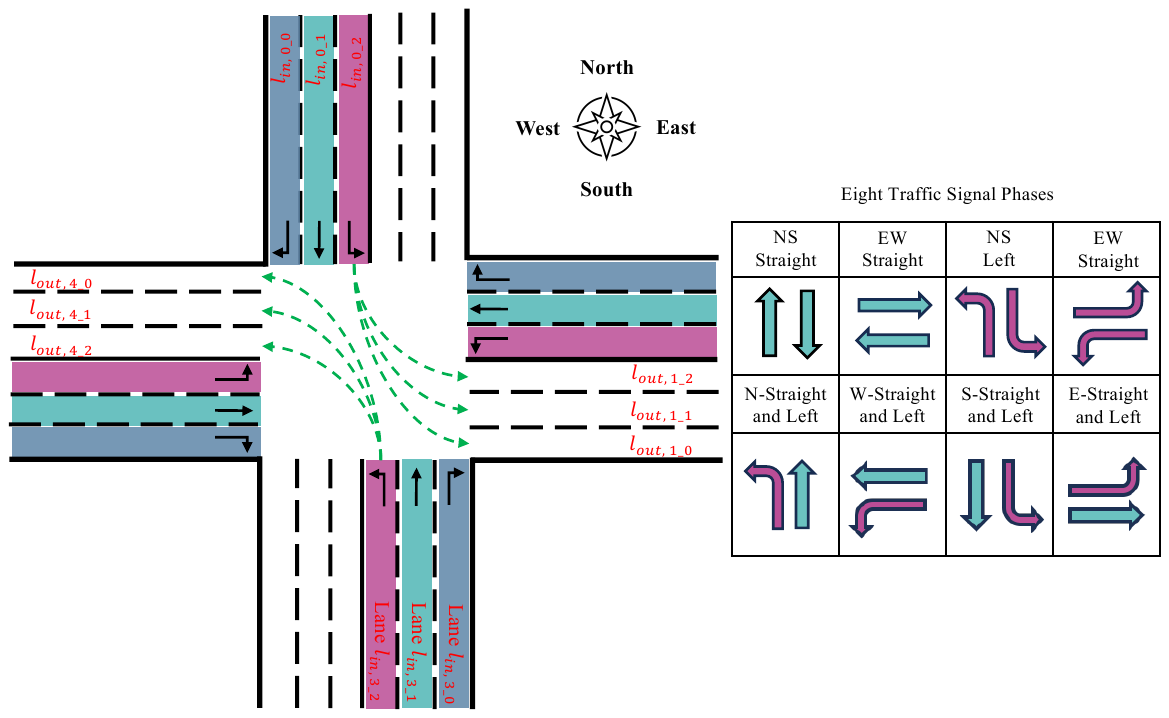}
\vspace{-0.7cm}
\caption{An illustration of a single intersection with eight traffic signal phases, where the phase $\textit{NS-Left}$ is currently activated, allowing the vehicles along the specified traffic movements (green dotted lines) to traverse the intersection.} 
\vspace{-0.4cm}
\label{intersection}
\end{figure}

\noindent \textbf{\textit{Definition 3 (Traffic signal phases)}}: An intersection features multiple traffic signal phases, where each signal phase $P$ is defined as a set of collision-free, simultaneously-authorized traffic movements $P \subseteq \mathcal{M}$, with $\mathcal{M}$ the set of all possible traffic movements of an intersection. 

\noindent\textbf{\textit{Definition 4 (Traffic Agent and Network)}}: 
A traffic agent optimizes traffic flows at an intersection by selecting phases and/or phase duration. 
A traffic network $\mathcal{G}(\mathcal{V},\mathcal{E})$ is structured as a multi-agent system with vertices $\mathcal{V}$ representing traffic agents and edges $\mathcal{E}$ signifying the inter-agent road connections.

\noindent \textbf{\textit{Definition 5 (Traffic Neighborhood)}}:
For any given agent $i \in \mathcal{V}$, its set of direct neighbors is denoted as $\mathcal{N}^{(1)}_{i} = \left\{j \mid j \in \mathcal{V} \wedge \textit{dist}(i, j) = 1 \right\}$, where $\textit{dist}(i, j)$ represents the shortest path measured in edges between agent $i$ and agent $j$. The local neighborhood $\mathcal{H}_{i}$, including both the agent itself and its direct neighbors, is defined as $\mathcal{H}_{i} = \{{i}\} \; \bigcup \; \mathcal{N}^{(1)}_{i}$.

Fig.~\ref{intersection} depicts an example single intersection composed of twelve incoming lanes and twelve outgoing lanes.
Thirty-six traffic movements are possible since each incoming lane can be connected to multiple outgoing lanes. Given the assumption that the right-turn lanes are never restricted, we formulate eight distinct (collision-free) phases, as illustrated on the right-hand side of Fig.~\ref{intersection} which includes \textit{NS-Straight}, \textit{WE-Straight}, \textit{NS-Left}, \textit{WE-Left}, \textit{S-Straight and Left}, \textit{W-Straight and Left}, \textit{N-Straight and Left}, and \textit{E-Straight and Left}. The intersection is currently operating under the $\textit{NS-Left}$ phase, allowing vehicles in the left-turn lanes from both North and South directions (green dotted lines) to cross the intersection.

\subsection{Multi Agent Reinforcement Learning}
In this section, we formulate the Multi-agent Traffic Signal Control (MATSC) problem as a Multi-Agent Reinforcement Learning (MARL) problem, where each intersection is represented as an independent learning agent. Accordingly, this problem can be modeled as a Decentralized Partially Observable Markov Decision Process (Dec-POMDP)~\cite{gupta2017cooperative}, which consists of a tuple $G = (N, S, A, R, T, O, Z, \rho, \gamma)$. 
Here $N$ represents the number of learning agents, $S$ the global traffic state space, and $\mathbf{A} = A_1 \times A_2 \times \ldots \times A_N$ the joint action space. 
The reward function $R: S \times \mathbf{A} \times S \rightarrow \mathbb{R}^n$ computes $N$ rewards $\mathbf{r}=[r_1, r_2, \cdots, r_N]$ at each time step, with each reward corresponding to one of the $N$ agents. 
The transition function $T: {S} \times \mathbf{A} \times {S} \to [0,1]$ determines the probability of transitioning from state $s$ to state $s'$ after executing the joint action $\mathbf{a}=[a_1, a_2, \ldots, a_N]$.  In a partially observable setting, agents do not have access to the the global traffic state $s$. Instead, each agent $i$ draws an individual observation $z_i$ from its observation function  $O_i(s): S \rightarrow Z_i$, where $Z_i$ represents the observation space of agent $i$. 
Finally, $\rho(s_0)$ and $\gamma$ represent the initial state distribution and discount factor respectively. 
Consider $\pi_i(a_i | z_i): Z_i \times A_i \to [0,1]$ as the stochastic policy for agent $i$, the joint policy for all agents is given by $\mathbf{\pi}(\mathbf{a}^t | \mathbf{z}^t) = \prod_{i=1}^{N} \pi_{i} (a^t_i|z^t_i)$. 
The objective is to find an optimal joint policy $\pi^{*}$ that can maximize the expected discounted return (i.e., discounted cumulative rewards) over all agents $J(\pi) = \mathop{E}_{\tau \sim \pi}[ \sum _{i=1}^{N} \sum_{t=0}^{\infty} \gamma^t \, r_i^{t}]$, where $\tau$ denotes a global state-action trajectory $(\mathbf{s}^0,\mathbf{a}^0,\mathbf{s}^1,\mathbf{a}^1, \cdots,\mathbf{s}^{t^e},\mathbf{a}^{t^e})$ with $t^e$ representing the trajectory length.

\subsection{RL Formulation}
We review the state, action, and reward choices for traffic signal control problems from prior work and then introduce our definitions used in the CoordLight framework.

\subsubsection{State}
At each time step, agents construct a local state of the traffic conditions at their intersections as a set of lane-specific features, collected from digital cameras or induction loop detectors. Prior work has used a combination of traffic phase, queue lengths, waiting time, delay, and pressure~\cite{chu2019multi,chen2020toward,wei2019presslight,wei2019colight}, as suitable local states. In CoordLight, we introduce a novel state called Queue Dynamics State Encoding (QDSE) derived from the queueing dynamics model. The details of the state representation QDSE are presented in Section.~\ref{method:QDSE}.

\subsubsection{Action}
 In MATSC, the action space comprises one or more of the following options: a) current phase duration given minimum and maximum duration constraints~\cite{aslani2017adaptive, aslani2018traffic}, b) cycle-based phase ratios~\cite{abdoos2011traffic, casas2017deep}, c) holding or changing the current phase within a predefined cycle sequence~\cite{brys2014distributed, mannion2016experimental, wei2018intellilight}, and d) choosing the subsequent phase without sequence to enable ATSC~\cite{wei2019presslight, wei2019colight, chen2020toward, zhang2022expression, lin2023denselight, goel2023sociallight}. In this work, we choose the action space as a finite collection of collision-free traffic phases. Agents will synchronously choose a traffic phase from this set (their action), and implement it for a predetermined duration (i.e., without a fixed cycle among phases), allowing for maximum adaptability. Note that this choice was made for fair comparison with prior methods; our method can adapt flexibly to other action spaces that meet the safety/fairness requirements of real applications.

\subsubsection{Reward}
Global reward objectives for MATSC optimize the average travel delay for all vehicles operating in the network. However, directly optimizing these global metrics in a decentralized manner has been proven to be challenging~\cite{zheng2019diagnosing}. Hence, prior work have optimized surrogate local reward structures such as queue length, waiting time, speed, accumulated delay, number of stops, traffic flow intensity, and pressure~\cite{chu2019multi,chen2020toward,wei2019presslight,wei2019colight}.
Among these options, queue length (the count of stopped vehicles) has been shown to yield robust and improved traffic performance~\cite{zheng2019diagnosing}.
Thus, we define the regional coordinated reward structure for each traffic agent as the negative sum of queue length on both incoming lanes and outgoing lanes, which can be written as:
$R(s,a)=-\left(\sum_{l_{in} \in \mathcal{L}_{in}} Q_{l_{in}} + \sum_{l_{out} \in \mathcal{L}_{out}} Q_{l_{out}} \right)$, where $ Q_{l_{in}}$ and $Q_{l_{out}}$ represents the incoming queue length and outgoing queue length, respectively.
Given that the ego agent's outgoing lanes are the incoming lanes of neighboring agents, this reward structure allows each agent to concurrently optimize its own regional objective (queue length of its incoming lanes) plus the objectives of its neighboring agents (queue length of its outgoing lanes, which are the connected incoming lanes of the neighbors) for enhanced regional cooperation.
Specifically, we believe that the impact of better local decision-making would spread over the entire network as local traffic transitions are highly coupled within a neighborhood.

\section{Queueing Dynamic State Encoding}
\label{method:QDSE}
\subsection{Vehicle Queueing Dynamics}
Efficient state representation in traffic signal control is crucial for RL agents to make high-quality decisions, e.g., to enable them to proactively anticipate how their actions (and those of their neighboring intersections) may influence their queue length.
This predictive capability shifts agents from merely reacting to traffic changes towards strategically optimizing traffic flow and proactively reducing congestion.
To this end, we first study the mechanism of vehicle queueing dynamics, and describe the evolution of the queue length as a discrete time process, where $t$ represents the time at which the agent decides/enacts a given phase, and $t+1$ marks the end of that phase (i.e., the next decision step, following a pre-determined phase duration).
The resulting discrete-time queue length dynamics for an isolated intersection reads:
\begin{equation}
\label{eq:vehicle_queueing_dynamics}
Q(t+1) = Q(t) + \Delta_{in}(t) - \Delta_{out}(t),
\end{equation}
\noindent where $Q(t)$ denotes the time-dependent queue length vector, and $\Delta_{in}(t)$ and $\Delta_{out}(t)$ are the incoming and outgoing queue length increment vectors (number of vehicles respectively added/removed to the queue during phase $P(t)$).
Each vector, expressed as $ Q(t) = [Q^l(t) \mid l \in \mathcal{L}_{in} ]$, $ \Delta_{in}(t) = [\Delta_{in}^l(t) \mid l \in \mathcal{L}_{in} ]$, and $ \Delta_{out}(t) = [\Delta_{out}^l(t) \mid l \in \mathcal{L}_{out} ]$, encompasses the respective vectors for all incoming lanes within the intersection. Note that the elements within these vectors are ordered according to the sequence of $l$ in $\mathcal{L}_{in}$.
Specifically, $\Delta_{out}^{l}(t) \in [0, C]$ if the traffic light is green on that lane during phase $P(t)$, with $C$ (throughput) the maximum number of vehicles that can pass during a green phase.

However, direct inference of $\Delta_{in}(t)$ is non-trivial, as this vector mainly represents the projected count of moving vehicles about to join the waiting queue before the subsequent decision step ($t+1$).
To accurately calculate/predict the count of these vehicles, we rely on the Intelligent Driver Model (IDM)~\cite{treiber2000congested}, a widely acknowledged car-following framework used to model the dynamics of individual vehicles and their responses to surrounding traffic conditions.
We denote the set of active/moving vehicles currently on the incoming lanes as $V_r(t)$.
We further define the distance between a vehicle and the far end of the queue (away from the intersection) as $D(P(t), Q(t))$ (a measured quantity, e.g., through cameras).
When the queue length is zero, $D(P(t), Q(t))$ is measured from the stop line of the intersection.
Given the instantaneous velocity $v_{i} (t)$ of vehicle (a measured quantity) at the beginning of phase $t$, and using the acceleration formulation $ \Dot{v}_i(t)$ from the IDM model, we can estimate the predicted distance $\widetilde{D}_i(t)$ that vehicle $i$ will travel during phase $t$:
\begin{equation}
\widetilde{D}_{i} (t) = \int_{t}^{t+1} \left(v_{i}(t) + \int_{t}^{t+t^{'}} \Dot{v}_i(t^{''}) dt^{''}\right) dt^{'}.
\label{eq:Dtilda}
\end{equation}
Using Eq.~\eqref{eq:Dtilda}, we can finally predict whether each moving vehicle will join the queue during phase $t$, by comparing their predicted travel distance $\widetilde{D}_{i}(t)$ to their actual distance from the end of the queue, $D(t)$.
We finally approximate $\Delta^{l}_{in}(t)$ of an incoming lane at the intersection as:
\begin{equation}
\Delta^{l}_{in}(t) = |\left\{i \mid i \in V^{l}_r(t) \; \wedge \; \widetilde{D}_{i}(t) \geq D_{i}(P(t), Q^{l}(t)) \right\}|.
\label{eq:delta_in}
\end{equation}
Note that Eq.~\eqref{eq:delta_in} technically underestimates the true value of $\Delta_{in}(t)$, as it ignores vehicles that will join the queue, as the queue length will increase; that is, Eq.~\eqref{eq:delta_in} simplifies the queue dynamic by assuming that $D_i(t)$ is constant during phase $P(t)$, which is at least slightly incorrect at times.
However, we note that the incurred error is small for short phases. 

In the context of traffic signal control problems, $\Delta_{out}(t)$ predominantly associates with the model's short-term efficacy, actively responding/reacting to real-time traffic conditions to alleviate congestion.
In contrast, $\Delta_{in}(t)$ is more aligned with the model's long-term capability, foreseeing/predicting impending congestion through an understanding of moving vehicle dynamics.
Hence, we argue in this work that a comprehensive and accurate depiction of queueing dynamics, especially in relation to predicting future congestion, can elevate the model's comprehension of congestion formulation, ultimately contributing to enhanced performance.

\subsection{State Representation} 
\label{sec:QDSE}
Given the complex nature of the vehicle queueing dynamics, traditional state definitions, such as vehicle count or pressure, fail to provide a comprehensive depiction of the local traffic conditions for RL agents to make informed decisions.
Therefore, we propose a novel traffic state definition, named Queue Dynamic State Encoding (QDSE) that encapsulates critical traffic feature variables inspired by queuing dynamics models to provide an enhanced insight into forthcoming congestion at a given intersection.
Specifically, QDSE is defined as a combination of six time-dependent lane-feature vectors, resulting in the state vector $S(t) \in \mathbb{R}^{6 \times |\mathcal{L}_{in}|}$ for each intersection:
\begin{equation}
S(t) = [Q(t), N_{\text{in}}(t), N_{\text{out}}(t), N_{\text{r}}(t), N_{\text{fr}}(t), D_{\text{fr}}(t)],
\end{equation}  
which includes the queue length, i.e., number of stopped vehicles $Q(t)$, entering vehicles $N_{\text{in}}(t)$, departing vehicles $N_{\text{out}}(t)$,  moving vehicles $N_{\text{r}}(t)$, the distance between the end of the halting queue and the foremost moving vehicle $D_{\text{fr}}(t)$, and the number of moving vehicles that follow the foremost moving vehicle within a specified distance $N_{\text{fr}}(t)$.

Our state representation QDSE includes queuing dynamics-specific features such as the current queue length $Q(t)$, which directly corresponds to the primary variable of the vehicle queueing dynamics equation, given in Eq.~\eqref{eq:vehicle_queueing_dynamics}. 
The features $N_{\text{in}}(t)$, $N_{\text{r}}(t)$, $N_{\text{fr}}(t)$, and $D_{\text{fr}}(t)$ capture the dynamics of approaching vehicles that may join and increase the queue, particularly those closest to the intersection, as they are strongly related to the queue length increment term $\Delta_{in}(t)$ in the vehicle queueing dynamics.
We believe that this may enhance the RL agents' ability to estimate $\widetilde{D}_{i} (t)$, the predicted travel distance of vehicles during the phase $t$, as detailed in Eq.~\eqref{eq:Dtilda}, thereby enabling more accurate representation of the incoming queue length $\Delta_{in}(t)$ term, as described in Eq.~\eqref{eq:delta_in}.
Meanwhile, $N_{\text{out}}(t)$ represents the number of vehicles leaving the intersection, which corresponds to $\Delta_{out}(t)$ and can be used by agents to identify the current phase $P(t)$.

\begin{figure}[t]
\centering
\includegraphics[width=8.2cm,height=2.2cm,trim=60 160 40 220, clip]{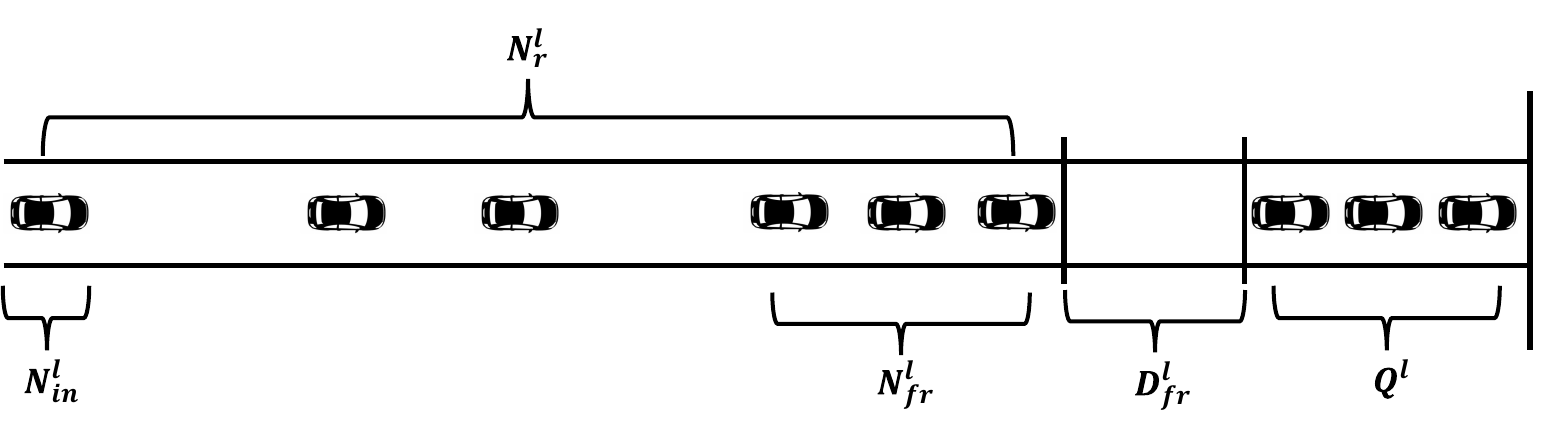}
\vspace{-0.6cm}
\caption{An illustration of the proposed state representation QDSE for an incoming lane $l$ of a single intersection, where $Q^l(t)=3 $, $N^l_{in}(t)=1$, $N^l_{out}(t)=0$,  $N^l_{fr}(t)=3$, $N^l_{r}(t)=6$, and $D^l_{fr}(t)=15$ in this case.}
\label{QDSE}
\vspace{-0.2cm}
\end{figure}

We believe that these specifics provide a more comprehensive depiction of the traffic state, extending beyond simple vehicle count or pressure that may allow the RL agent to reason about the current and upcoming traffic conditions in a more nuanced manner. 
When agents are equipped with a comprehensive and predictive view of traffic dynamics, they can synchronize their actions more effectively, thus enhancing coordination across the network. 
Additionally, the integration of multiple lane features in QDSE enhances the robustness of the learned policies, allowing them to adapt to complex and varying traffic flows. 
By focusing on specific lane features rather than detailed speed and position for each vehicle, QDSE strikes a balance between the complexity of state representation (e.g., image-based methods, which offer high accuracy but also high complexity) and accuracy (e.g., simple vehicle counts, which reduce complexity but at the cost of precision), thereby enhancing its applicability and effectiveness toward real-world ATSC applications.
Fig.~\ref{QDSE} offers a detailed illustration of the proposed state definition in an example case for an incoming lane at the intersection.

\section{Neighbor-aware Policy Learning}
\label{method:NAPO}
\subsection{Independent Policy Optimization}
To solve for the decentralized MATSC problem, a simple solution is to rely on independent learning~\cite{de2020independent, chen2017decentralized}, where each agent (intersection) learns to optimize its own policy function $\pi_{\theta_i}(z_{i}^{t})$, parameterized by $\theta_i$ to maximize its cumulative individual rewards. 
The observation of agent $i$ at decision time step $t$ is denoted as $z_i^t = (s_i^t, s_{\mathcal{N}_i}^{t})$, which integrates its local traffic state $s_i$ with the states of neighboring agents $s_{\mathcal{N}_i}=\left\{s_j \mid j \in \mathcal{N}_i\right\}$ via local communications.
Additionally, each agent learns a local value function parameterized by $\phi_i$ to estimate the expected cumulative rewards from a given state over time, expressed as $V_{\phi_i}(z_{i}^{t})=\mathbb{E}_{\tau}[\sum_{t'=t}^{t_e} \gamma^{(t'-t)} \; r_i^t]$.
Given the advantage function as $A_{i}^{t}(z_i^t, a_i^t)= r_{i}^{t} + \gamma \; {V}_{\phi_i}(z_i^{t+1}) - {V}_{\phi_i}(z_i^t)$, the policy gradient-based methods~\cite{mnih2016asynchronous} compute the gradient for parameter updates as follows:
\begin{equation}
\nabla_{\theta_i} J\left(\theta_i\right)={\mathbb{E}}_{\tau}\left[\nabla_{\theta_i} \log \pi_{\theta_i}\left(a_i \mid z_i \right) A_i \left(z_i , a_i\right)\right].
\end{equation}
Furthermore, parameter sharing is often applied in independent learning to improve data efficiency and tackle non-stationarity in the learning process within complex multi-agent environments, as demonstrated in studies such as~\cite{mappo, goel2023sociallight, de2020independent, damani2021primal}. Through this technique, agents synchronously update a commonly shared policy network and value network, with the parameters represented as $\theta=\theta_i=\ldots=\theta_N$ for the policy network and $\phi=\phi_i=\ldots=\phi_N$ for the value network.
However, such independent learning algorithms often lead to sub-optimal solutions because the learning agents optimize their individual rewards in a greedy manner, treating other agents as part of the environment dynamics. Consequently, these methods tend to overlook the crucial interactions and potential inter-dependencies among agents, which impedes the learning of efficient coordination and cooperation. 

\subsection{Neighbor-aware Actor-Critic Network}

Given the dynamics of a traffic network, an agent (junction) needs to have both spatial and temporal awareness of its surroundings to make informed decisions, e.g., to deduce the movement status of moving vehicles without full knowledge of their underlying dynamics/intentions. 
Since traffic agents are significantly influenced by their adjacent agents, and these impacts dynamically change over time, identifying and analyzing the dynamics of influential neighbors is crucial for effectively managing traffic flows and achieving enhanced collective traffic optimization.
Therefore, we introduce neighbor-aware learning, wherein we propose a policy function that learns to adaptively weight the importance of neighbors' states and integrate relevant historical information to enhance the decision-making process. 
Moreover, we present a privileged value function that aggregates critical historical state-driven action awareness of neighbors, aiming to facilitate more accurate value estimation to enhance policy updates and stabilize the training process.
The neighbor-aware policy function and value function can be formally expressed as:
\begin{equation}
    \hat{\pi}_{i}(t) = \hat{\pi}_{\theta}(\hat{z}_i^{\tau(:t)}), \;
    \hat{V}_{i}(t) = \hat{V}_{\phi}(\hat{z}_i^{\tau(:t)}, \hat{a}_{\mathcal{N}_{i}}^{\tau(:t)}),
\end{equation} 

\begin{equation}
    \hat{z}_i^t= (s_{i}^{t}, \alpha_{i}^t \; s_{\mathcal{N}_{i}}^{t}), \;
    \hat{a}_{\mathcal{N}_{i}}^{t}= (\beta_{i}^t \; a_{\mathcal{N}_{i}}^{t}),
\end{equation}

\begin{equation}
    \mathbf{\alpha}_{i}^t = \mathcal{F}_{\xi}(s_i^{t}, s_{ \mathcal{N}_{i}}^{t}),\;
    \mathbf{\beta}_{i}^t = \mathcal{F}_{\psi}(\hat{z}_i^{t}, a_{\mathcal{N}_{i}}^{t}),
\end{equation}
where $\hat{z}_i^{\tau(:t)}$ and $\hat{a}_{\mathcal{N}_i}^{\tau(:t)}$ are the augmented observation sequence and action sequence of agent $i$, respectively, across time.
Here, $(\cdot)^{\tau(:t)}$ represents the operation that processes a time sequence of input vectors from initial step to the decision step $t$.
Specifically, given the state of the ego agent and its neighbors at each decision step $t$, the augmented observation vector is calculated as: $\hat{z}_i^t= (s_{i}^{t}, \mathbf{\alpha}_{i}^t \, s_{\mathcal{N}_{i}}^{t})$, where $\mathbf{\alpha}_{i}^t  \in \mathbb{R}^{|\mathcal{N}_i|}$ is a learnable vector parameterized by $\xi$, used to evaluate the importance of the neighbors' joint state $s_{\mathcal{N}_i}^{t}$ given the current ego local state $s_i^{t}$.
$\mathbf{\beta}^t  \in \mathbb{R}^{|\mathcal{N}_i|}$ is another learnable vector parameterized by $\psi$ that learns to dynamically estimate the impacts of neighbors' current actions $\mathbf{a}_{\mathcal{N}_{i}}^{t}$ given the ego augmented observation $\hat{z}_i^{t}$.
These neighbor-aware vectors enable agents to reason about the crucial influence caused by their neighbors. Recognizing these influences separately, rather than as a fused whole, ensures a clearer representation of individual behaviors and contributions of neighboring agents.

\begin{figure*}[t]
\centering
\includegraphics[width=\textwidth, trim=0cm 0cm 0cm 0cm, clip]{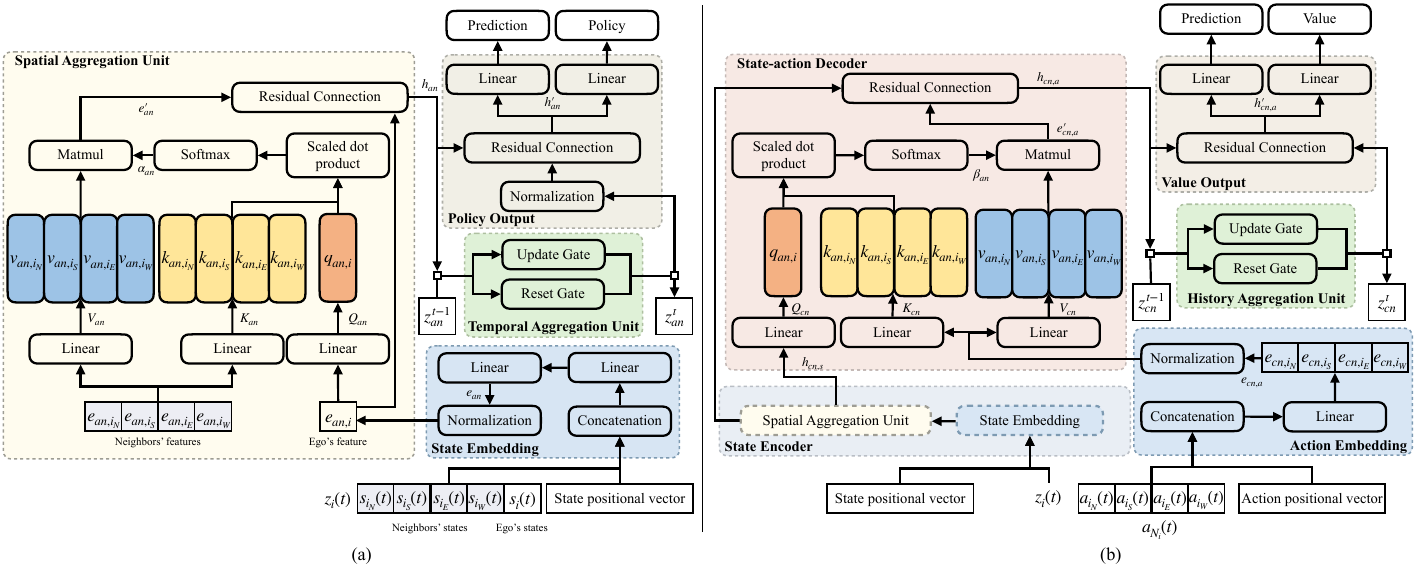}
\vspace{-0.8cm}
\caption{Detailed structure of our neighbor-aware actor-critic network in CoordLight: Fig. (a) shows the overall structure of our attention-based, spatio-temporal actor network (STN). This network consists of a spatial aggregation unit and a temporal aggregation unit to process information from the agent's neighborhood for policy output. Fig.(b) illustrates the structure of the privileged local critic network, which includes a state encoder and a state-action decoder that captures crucial interactions and incorporates neighbors' state-action dependencies into the value estimation.
}
\vspace{-0.3cm}
\label{network}
\end{figure*}

We further design an attention-based actor-critic network to implement neighbor-aware learning, which is detailed below.
\subsubsection{Actor Network}
First, we concatenate the observation vector $z_i(t) = (s_i^t, s_{i_N}^t, s_{i_S}^t, s_{i_E}^t, s_{i_W}^t)$ (where $i_N, i_S, i_E, i_W$ represent the neighbors of $i$ to the North, South, East and West, respectively) with a state positional vector (a one-hot vector that identifies agent indices) as the input vector. 
The input is projected into a higher-dimensional feature vector $e_{an} \in \mathbb{R}^{5 \times d}$ through multiple linear layers. This vector is then fed to the spatial aggregation unit, where we utilize a multi-head attention mechanism~\cite{vaswani2017attention} to aggregate neighbors' state information. Specifically, for each attention head, the feature vector is divided, and processed through three different linear layers (with the same hidden dimension $d$) to generate the $\mathbf{Q}$ (query), $\mathbf{K}$(key), and $\mathbf{V}$(value) vectors of the attention mechanism. 
Here, $\mathbf{Q}_{an} \in \mathbb{R}^{1 \times d} = [q_{i}]$ is a single query vector calculated from the normalized ego feature vector. 
Similarly, the key and value vectors, denoted as $\mathbf{K}_{an} \in \mathbb{R}^{4 \times d} = [k_{an,i_N}, k_{an,i_S}, k_{an,i_E}, k_{an,i_W}]$ and $\mathbf{V}_{an} \in \mathbb{R}^{4 \times d} =[v_{an,i_N}, v_{an,i_S}, v_{an,i_E}, v_{an,i_W}]$ respectively, are derived from the normalized neighbor feature vector. 
Next, the attention weight vector $\alpha_{an}$ is computed based on the scaled-dot product between the key vector and query vector: $\alpha_{an} =\operatorname{softmax}\left({\mathbf{Q}_{an}} \cdot \mathbf{K}_{an}^{T}) / \sqrt{d} \right)$.
Finally, we calculate the output vector $e^{'}_{an}$ as the weighted sum of all the value vectors, using these learned attention weights: $e^{'}_{an} \in \mathbb{R}^{1 \times d} = \alpha_{an} \: \mathbf{V}_{an}$.
The resulting vectors from different heads are concatenated and then combined with the ego feature vector using a residual connection to yield the state-aggregated feature vector $h_{an}$.
We further employ a recurrent neural network (specifically, a Gated Recurrent Unit, GRU~\cite{chung2014empirical}) to automatically aggregate crucial state features temporally.
Finally, we obtain a policy by projecting this feature vector $h^{'}_{an}$ through a linear layer followed by a softmax activation function. In parallel, the feature vector $h^{'}_{an}$ is fed through an additional linear layer to predict the queue length vector at the subsequent decision step. 
This branch aims to enhance the model's ability to represent queue length transition dynamics via supervised learning. 
The specifics of the actor network can be found in Fig.~\ref{network}(a).

\subsubsection{Critic Network}
The privileged local critic network comprises two components: a state encoder and a state-action decoder. Its detailed structure is shown in the Fig.~\ref{network}(b), with descriptions for each component provided as follows:

\textbf{State Encoder}:
The state encoder employs the same state embedding module and spatial aggregation unit used in the actor network, to transform the observation $z_i(t)$ with a state positional vector into a higher dimensional feature vector. The derived hidden feature vector $h_{cn,s}$ is then used as the query vector $\mathbf{Q}_{cn} \in \mathbb{R}^{1 \times d}$ in the state-action decoder.

\textbf{State-action Decoder}: 
We embed and transform the current neighbors' action vector $a_{\mathcal{N}_i}(t)=[a_{i_{N}}^t, a_{i_{S}}^t, a_{i_{E}}^t, a_{i_{W}}^t]$, concatenated with the action positional vector (a one-hot vector used to denote the indices of four neighboring agents) into a higher-dimensional vector $e_{cn,a} \in \mathbb{R}^{4 \times d}$.
Then, we calculate the attention weights of neighbors' current actions given the ego augmented observation feature vector through the same multi-head attention mechanism, wherein the query vector $\mathbf{Q}_{cn} \in \mathbb{R}^{1 \times d} = [q_{cn, i}]$ is derived from the normalized aggregated feature vector $h_{cn, s}$ of the state encoder.
The key vector, $\mathbf{K}_{cn} \in \mathbb{R}^{4 \times d} = [k_{cn, i_N}, k_{cn, i_S}, k_{cn, i_E}, k_{cn, i_W}]$ and value vector, $\mathbf{V}_{cn} \in \mathbb{R}^{4 \times d} = [v_{cn, i_N}, v_{cn, i_S}, v_{cn, i_E}, v_{cn, i_W}]$, are obtained after normalizing the action feature vector $e_{cn, a}$. 
Thus, the attention weights for each head, using the scaled-dot product, can be calculated as:
$\beta_{cn} = \operatorname{softmax}\left((\mathbf{Q}_{cn} \cdot \mathbf{K}_{cn}^{T}) / \sqrt{d} \right)$.
We further calculate the output vector as the weighted sum of the value vector using the computed attention weights: $e_{cn,a}^{'} \in \mathbb{R}^{1 \times d} = \beta_{cn} \, \mathbf{V}_{cn}$. These vectors from different heads are concatenated and then added with the original feature vector $h_{cn,s}$ from the state encoder through a residual connection to obtain the feature vector $h_{cn,a}$. Additionally, we pass $h_{cn,a}$ through another GRU module to selectively aggregate the historical augmented neighbor-aware state-action information. Finally, the resulted feature vector $h^{'}_{cn,a}$ is used to output the state-value estimate and the queue length prediction vector through two different linear layers.
In cases of missing or incomplete neighbor information, we pad the missing data with zeros and apply a mask in the attention mechanism. This ensures consistent input dimensions and prevents agents from focusing on incomplete or incorrect neighbor information.

\subsection{Neighbor-aware Policy Optimization}
We enhance the independent learning algorithm, \textit{Independent Proximal Policy Optimization} (IPPO)~\cite{de2020independent, schulman2017proximal} with our neighbor-aware learning to propose our fully decentralized MARL algorithm, termed \textit{Neighbor-aware Policy Optimization} (NAPO).
This algorithm aims to mitigate the agents' myopic behaviors and promote the learning of efficient coordination strategies by deepening the understanding of the potential influence and interactions of each neighbor.
Within NAPO, we first incorporate the neighbor-aware value function into the advantage calculation, the resulted neighbor-aware advantage function thus can be formally expressed as:
\begin{equation}
\begin{gathered}
\hat{A}_{i}^{t}(z_i^t, a_i^t) = r_{i}^{t} + \gamma \; \hat{V}_{i}^{t+1}  - \hat{V}_{i}^{t} \\
= r_{i}^{t} + \gamma \; \hat{V}_{\phi}(\hat{z}_i^{\tau(:t+1)}, 
\hat{{a}}_{\mathcal{N}_{i}}^{\tau(:t+1)})  - \hat{V}_{\phi}(\hat{z}_i^{\tau(:t)}, \hat{a}_{\mathcal{N}_{i}}^{\tau(:t)}).  
\end{gathered}
\end{equation}
In this advantage function, the parametric vectors $\alpha_{\xi}$ and $\beta_{\phi}$ serve as two adaptive weighting factors that prioritize the importance of neighboring agents' states and actions respectively by assigning different weights at different situations. Thereby, this neighbor-aware advantage function can effectively capture the dependencies between adjacent agents, i.e., focusing on influential neighbors by assigning higher weights and ignoring irrelevant neighbors by assigning lower weights, thus leading to a more accurate estimation of the contribution of ego actions to enhance coordinated policy learning. 
According to~\cite{coma}, the baseline $\hat{V}_{\phi}$ used in the neighbor-aware advantage function is not conditioned on the ego agent's actions. 
This ensures that it does not introduce any bias into the policy gradient estimator, thereby guaranteeing the convergence of NAPO.
Combined with the General Advantage Estimate (GAE) method~\cite{schulman2015high}, the final advantage function can be written as:
\begin{equation}
\label{eqn:gae}
    \bar{A}_{i}^{t}={A}^{GAE}({z}_{i}^{t},a_i^t) = \sum_{l=0}^{T} (\gamma \lambda)^l \hat{A}_i^{t+l},
\end{equation}
where $\lambda$ controls the trade-off between bias and variance. 
Thus, the policy loss of NAPO, used to update the policy function (i.e., the actor network), for a single agent $i$ over a given trajectory of length of $T$, is defined as: 
\begin{equation}
\label{eqn:policy_loss}
L^{(i)}(\theta) = - \;\frac{1}{T} \sum_{i=0}^{T} \; \min \left(\hat{s}_{\theta}^{t} \, \bar{A}_i^{t}, \operatorname{clip}\left(\hat{s}_{\theta}^{t}, 1-\epsilon, 1+\epsilon\right)\, \bar{A}_i^{t}\right),
\end{equation}
where $ \hat{s}_{\theta}^{t} = \frac{\hat{\pi}_\theta\left(a_i^{t} \mid \hat{z}_i^{\tau(:t)}\right)}{\hat{\pi}_{\theta_{old}}\left(a_i^{t} \mid \hat{z}_i^{\tau(:t)}\right)}$ is the importance sampling factor which computes the discrepancy between current policy and previous policy to prevent drastic deviations during policy update. To enhance exploration and avoid the policy to be too deterministic during the training phase, we integrate an additional entropy loss, which is defined as:
\begin{equation}
\label{eqn:entropy_loss}
    L_{e}^{(i)}(\theta) = \frac{1}{T} \sum_{t=0}^{T} \, \sum_{a_i \in A_i} \hat{\pi}_\theta(a_i^t \mid \hat{z}_{i}^{\tau(:t)}) \log \hat{\pi}_\theta(a_i^t \mid \hat{z}_i^{\tau(:t)}).
\end{equation} 

\begin{algorithm}[t]
\label{algorithm:napo}
\caption{Neighbor-aware Policy Optimization}
\begin{algorithmic}[1]
\STATE Initialize $\theta$, the parameters for policy function (actor) $\hat{\pi}$, and $\phi$, the parameter for value function (critic) $\hat{V}$
\WHILE{$episode < episode_{max}$}
	\STATE Initialize trajectory buffer $\tau$
 
	\STATE Initialize $ h_{0,\pi}^{(0)}, ..., h_{n,\pi}^{(0)}$, the hidden states for actor GRU
 
	\STATE Initialize $ h_{0,v}^{(0)}, ..., h_{n, v}^{(0)}$, the hidden states for critic GRU
 
    \STATE \texttt{// Collect trajectory data}
	
    \FOR{$t \leftarrow 1$ \TO $T$}
		\FOR{all agents $i$}
			\STATE $a_{i}^{(t)},\ h_{i,\pi}^{(t)} = \hat{\pi}(z_{i}^{(t)},\ h_{i, \pi}^{(t-1)}; \theta)$
		\ENDFOR
		\FOR{all agents $i$}
			\STATE Calculate neighbors' current actions, $a_{N_{i}}^{(t)}$
			\STATE $\hat{V}_{i}^{(t)},\ h_{i, v}^{(t)} = \hat{V}(z_{i}^{(t)},\ a_{N_{i}}^{(t)},\ h_{i, v}^{(t-1)}; \phi)$
		\ENDFOR
		\STATE Execute actions $\mathbf{a}^{(t)}$, observe $\mathbf{r}^{(t)}$, $\mathbf{z}^{(t+1)}$, $\mathbf{q}^{(t+1)}$
		\STATE Store $[\mathbf{z}^{(t)}, 
        \mathbf{\pi}^{(t)},
        \mathbf{a}^{(t)}, \mathbf{a}^{(t)}_{\mathcal{N}},
        \mathbf{r}^{(t)}, \mathbf{z}^{(t+1)}, \mathbf{q}^{(t+1)}]$ in $\tau$
	\ENDFOR
        \STATE Compute advantage estimate $\bar{A}$ via GAE (Eq.~\eqref{eqn:gae})
        \STATE Compute TD targets for critic network (Eq.~\eqref{eqn:value_loss})
        \STATE \texttt{// Update using PPO algorithm}
        \FOR{$epoch \leftarrow 1$ \TO $K$}
        \STATE Reset the hidden states for the actor GRU 
        \STATE Reset the hidden states for the critic GRU
		\FOR{all agents $i$}
                \STATE Compute new policy $\hat{\pi}_{i, new}$ using data $\tau$
                \STATE Compute queue length prediction $p_{\theta,i}$ and $p_{\phi,i}$ through the actor and critic network using data $\tau$
		\ENDFOR
		\STATE Update $\theta$ on ${L}(\theta)$ via Adam with data $\tau$ (Eq.~\eqref{eqn:actor_loss})
		\STATE Update $\phi$ on ${L}(\phi)$ via Adam with data $\tau$ (Eq.~\eqref{eqn:critic_loss})
	\ENDFOR
\ENDWHILE
\end{algorithmic}
\end{algorithm}

The critic network aims to approximate the state-value estimates by minimizing the Mean Squared Error (MSE). Defining the critic target as the Temporal Difference(TD) error, the value loss of NAPO is expressed as follows:
\begin{equation}
\label{eqn:value_loss}
L^{(i)}(\phi) = \frac{1}{T} \sum_{t=0}^{T} \left( \underbrace{r_i^t + \gamma \, \hat{V}_{i,\phi}^{t+1}}_{\text{TD Target}} - \hat{V}_{i,\phi}^{t}\right)^2.
\end{equation}
Moreover, we derive two vectors, $p_{\theta}$ and $p_{\phi}$ from the same actor network and critic network respectively to predict queue length on the incoming lanes and outgoing lanes at next decision step.
Given the ground of the truth of queue length vector at the next step as $\hat{q}_{i}^{t+1}$, the prediction losses for both actor network and critic network can be written as: 
\begin{equation}
\label{eqn:prediction_loss}
\begin{aligned}
    &L_{p}^{(i)}(\theta)= \frac{1}{T} \sum_{t=0}^{T} \left(p_{\theta, i}^{t}-\hat{q}_{i}^{t+1}\right)^2,\\
    &L_{p}^{(i)}(\phi)= \frac{1}{T} \sum_{t=0}^{T}  \left(p_{\phi, i}^{t} - \hat{q}_{i}^{t+1}\right)^2 .
\end{aligned}
\end{equation} 
Note that the output prediction is not employed for any decision-making processes; rather, our objective is to enhance the model's capability to represent future queueing dynamics by incorporating such auxiliary prediction loss.
The final loss functions to be minimized for updating the actor network and critic network over all $N$ learning agents are represented as:
\begin{equation}
\label{eqn:actor_loss}
{L}(\theta)= \frac{1}{N} \sum_{i=0}^{N} \; (L^{(i)}(\theta) + \omega_{e} \, L_{e}^{(i)}(\theta) + \omega_{p} \, L_{p}^{(i)}(\theta)),
\end{equation}
\begin{equation}
\label{eqn:critic_loss}
{L}_{(\phi)}=  \frac{1}{N} \sum_{i=0}^{N} \; (L^{(i)}(\phi) + \omega_{p} \, L_{p}^{(i)}(\phi)),    
\end{equation}
where $\omega_{e}$ and $\omega_{p}$ are two constant weighting coefficients to adjust the entropy loss and prediction loss respectively. 
Specifically, the neighbor-aware vectors $\alpha$ and $\beta$ are learned via the same RL losses in this work.  
It is worth noting that these vectors are also flexible to be learned via other specific loss functions to achieve different objectives. The details of our NAPO algorithm are illustrated in Algorithm.~\ref{algorithm:napo}.

\section{Experiments}
\label{experiments}

\begin{figure*}[t]
    \centering
    \includegraphics[width=\linewidth]{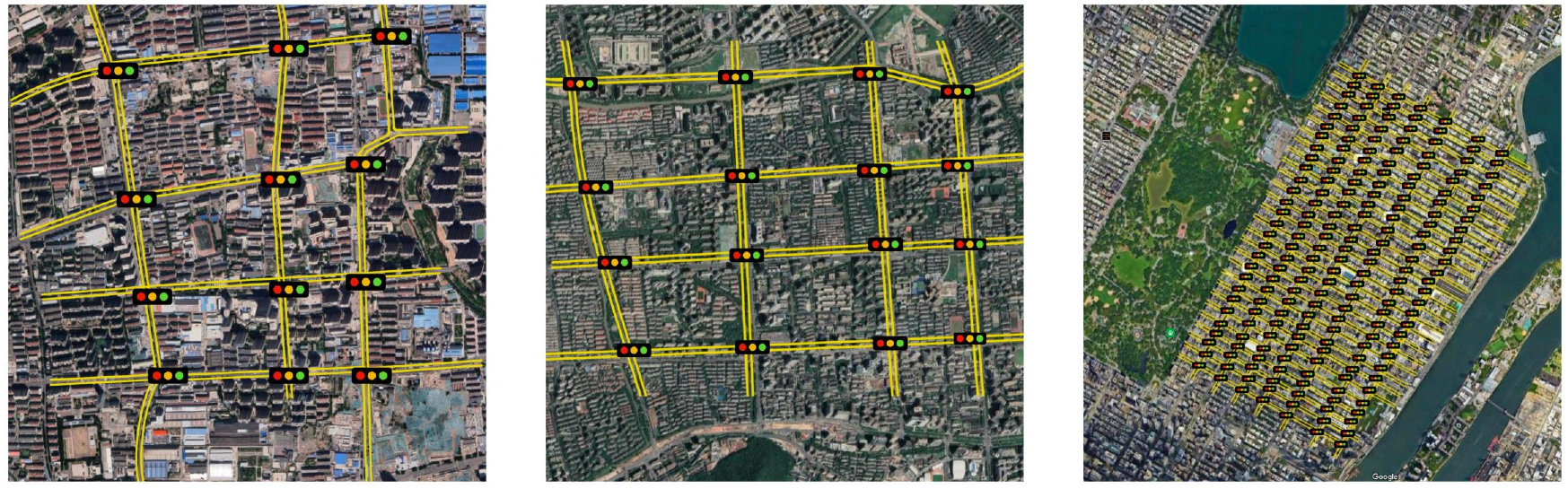}
    \vspace{-0.8cm}
    \caption{Real-world traffic road networks of the CityFlow simulation environments used in our experimental results: Jinan, China map (left, $3 \times 4$ intersections), Hangzhou, China map (center, $4 \times 4$ intersections), and New York, USA map (right, $7 \times 28$ intersections).}
    \label{maps}
    \vspace{-0.4cm}
\end{figure*}

Our experiments rely on the traffic maps and traffic demand datasets provided by~\cite{wei2018intellilight, wei2019colight, chen2020toward}. The road networks used for simulations are extracted from real-world traffic maps. Our training and testing episodes are capped at $3600$ seconds, and individual phases (decisions) are fixed to 5 seconds. At a decision step, if an agent (intersection) selects the same already-enacted phase, this phase is extended by 5 seconds. On the other hand, any phase transition is always preceded by 2 second of yellow lights (for all previously-green lights), resulting in a shorter, 3-second new phase (totalling 5 seconds).
We train a homogeneous traffic policy, i.e., the same policy weights for all intersection in each map and traffic demand, consistent with recent works in the field. We then evaluate CoordLight using the open-source traffic simulator - CityFlow~\cite{tang2019cityflow}, and benchmark it against both conventional and learning-based baselines using the average travel time as the primary global metric (smaller is better). 
Specifically, the average travel time is calculated as: $\frac{1}{N_v} \sum_{v=1}^{N_v} (t_{end}^{v}-t_{start}^{v})$, where $N_v$ is the total number of vehicles in the traffic network, $t_{start}^{t}$ and $t_{end}^{v}$ are the time that a vehicle enters and exists the traffic network.
Note that $t_{end}^{v}$ is set to $3600$ if the vehicle is unable to reach its destination before the end of simulation.
We also conduct additional ablation studies to assess the impact of our proposed state representation - QDSE and the key components of our MARL algorithm NAPO.

All experiments are run on an Ubuntu workstation with an AMD Ryzen 9 5950x, 32GB RAM, and an NVIDIA GeForce RTX 3060. 
For the training hyper-parameters, we establish a batch size of $720$, a learning rate of $0.0003$ for the actor network, and a learning rate of $0.0005$ for the critic network.
The loss coefficients are determined as follows: value loss at $0.5$, entropy loss at $0.01$, and prediction loss at $0.005$. Additionally, we set the discount factor and GAE factor both at $0.98$, a policy clip ratio of $0.2$, and $6$ epochs for PPO update. 
We set the hidden dimensions of the linear, attention, and GRU layers of the neural networks to 128 and adopt the Adam optimizer~\cite{kingma2014adam} for gradient descent and parameter updates. 

\subsection{Traffic Datasets}
The employed real traffic dataset includes three urban networks: Jinan, Hangzhou, and New York, key benchmarks for evaluating extensive ATSC methods~\cite{zheng2019frap, wei2019colight}.
The road networks are extracted from real-world traffic maps and uniformly feature homogeneous and regular intersections. Each intersection has four incoming roads, each comprising three lanes. The detailed configuration of the road networks is further illustrated in Fig.~\ref{maps}.
The traffic flow datasets are compiled from camera observations over various time periods, where a single road network may contain multiple flow datasets, each reflecting distinct traffic demand levels.
The Jinan map is composed of 12 intersections (3 $\times$ 4), Hangzhou of 16 intersections (4 $\times$ 4), and New York consists of 196 intersections (28 $\times$ 7).
The dataset comprises three traffic flow datasets for Jinan, two for Hangzhou, and two for New York. The volumes and arrival rates of different traffic demands of three city networks are summarized in Table.~\ref{dataset}.

\begin{table}[t]
\centering
\caption{Traffic dataset for three cities - Jinan, Hangzhou, and New York, each with different traffic demands.}
\vspace{-0.3cm}
\resizebox{\linewidth}{!}{%
\tiny
\begin{tabular}{ccccccc}
\hline
\multirow{2}{*}{Network} & \multirow{2}{*}{Type} & \multicolumn{1}{l}{\multirow{2}{*}{Volume (veh)}} & \multicolumn{4}{c}{Arrival Rate(veh/min)} \\ \cline{4-7} 
 &  & \multicolumn{1}{l}{} & Mean & Std. & Max & Min \\ \hline
\multirow{3}{*}{Jinan (3x4)} & $\mathcal{D}_{\text{JN(1)}}$ & 6295 & 104.92 & 19.79 & 136.00 & 50.00 \\
 & $\mathcal{D}_{\text{JN(2)}}$ & 4365 & 72.75 & 15.15 & 101.00 & 43.00 \\
 & $\mathcal{D}_{\text{JN(3)}}$ & 5494 & 91.57 & 9.51 & 111.00 & 69.00 \\ \hline
\multirow{2}{*}{Hangzhou (4x4)} & $\mathcal{D}_{\text{HZ(1)}}$ & 2983 & 49.72 & 8.24 & 67.00 & 40.00 \\
 & $\mathcal{D}_{\text{HZ(2)}}$& 6984 & 116.40 & 63.72 & 230.00 & 39.00 \\ \hline
\multirow{2}{*}{New York (7x28)} & $\mathcal{D}_{\text{NY(1)}}$ & 10676 & 177.92 & 27.06 & 222.00 & 79.00 \\
 & $\mathcal{D}_{\text{NY(2)}}$ & 15862 & \multicolumn{1}{l}{264.35} & \multicolumn{1}{l}{35.10} & \multicolumn{1}{l}{320.00} & \multicolumn{1}{l}{79.00} \\ \hline
\end{tabular}%
}
\label{dataset}
\vspace{-0.3cm}
\end{table}

\subsection{Baseline methods}

We benchmark the policies learned by CoordLight against diverse conventional and learning-based baseline methods.
Aligned with recent works in the field, for all learning-based methods, we rigorously train a specific policy for each traffic demand within every individual road network.
We evaluate the policies on 10 evaluation episodes with different seeds after each training and report the average value of vehicle travel time.
Note that we ran most of the baselines under our setup, but a few results could not be obtained. These results are reported from their original paper, and indicated by a $*$.

\subsubsection{Conventional Baselines}
\begin{enumerate}
    \item \textbf{FixedTime~\cite{koonce2008traffic}}: Employs fixed time intervals to each phase based on pre-defined cycle length and phase split.
    \item \textbf{MaxPressure(MP)~\cite{varaiya2013max}}: Aims to alleviate congestion by balancing traffic flows between upstream and downstream roads. It adopts a greedy strategy to choose the traffic phase that minimizes intersection pressure, which is calculated as the total difference in vehicle counts between incoming and connected outgoing lanes. 
    \item \textbf{Advanced-MP~\cite{zhang2022expression}}: An enhanced version of MP, which controls the phase selection based on the highest vehicle demand. It considers both moving and queueing vehicles within an effective range (determined by the maximum lane speed and phase duration) at intersections.
    
\end{enumerate}

\subsubsection{MARL Baselines}
\begin{enumerate}
    \item \textbf{CoLight~\cite{wei2019colight}} : A coordination method that employs Graph Attention Neural Networks (GATs) for junction-level cooperation and is trained via Deep Q-learning.
    \item \textbf{MPLight~\cite{chen2020toward}}: 
    Integrates pressure metrics into state and reward definitions, and employs a FRAP-based training architecture~\cite{zheng2019frap} that involves phase competition to enhance traffic control performance.
    \item \textbf{Advanced-CoLight, Advanced-MPLight~\cite{zhang2022expression}}: 
    Enhances CoLight and MPLight by incorporating the current phase and Advanced Traffic States (ATS), including efficient pressure and the count of vehicles within an effective range, into the intersection observation.
    \item \textbf{DenseLight~\cite{lin2023denselight}}:
    Proposes an unbiased reward function based on the ideal distance gap, paired with an enhanced feature extraction mechanism for non-local traffic conditions to achieve refined traffic signal control.
    \item \textbf{SocialLight~\cite{goel2023sociallight}}: 
     Facilitates scalable cooperation among junctions by distributedly estimating individual contributions within their local neighborhoods.
\end{enumerate}

\begin{table*}[]
\centering
\caption{Average travel times (sec) comparing CoordLight with baselines in each CityFlow environment/dataset ($*$ indicate results that had to be reported from the original baseline paper).}
\vspace{-0.3cm}
\resizebox{\textwidth}{!}{%
\tiny
\begin{tabular}{cccccccc}
\hline
Method            & $\mathcal{D}_{\text{JN(1)}}$    & $\mathcal{D}_{\text{JN(2)}}$    & $\mathcal{D}_{\text{JN(3)}}$    & $\mathcal{D}_{\text{HZ(1)}}$    & $\mathcal{D}_{\text{HZ(2)}}$    & $\mathcal{D}_{\text{NY(1)}}$ & $\mathcal{D}_{\text{NY(2)}}$     \\ \hline
Fixed-Time        & 346.36 & 289.74 & 316.70 & 432.32 & 359.44 & 1507.12                 & 1660.29 \\
MaxPressure       & 273.96 & 245.38 & 245.81 & 288.54 & 348.98 & 1179.55                 & 1535.77 \\
Advanced-MP       & 253.61 & 238.62 & 235.21 & 279.47 & 318.67 & 1060.41 $*$             & -       \\ \hline
CoLight           
& 276.33 $\pm$ 3.98 
& 237.14 $\pm$ 0.73 
& 278.16 $\pm$ 1.13 
& 271.07 $\pm$ 2.11 
& 297.26 $\pm$ 9.47 
& 1221.77 $\pm$ 19.83                
& 1476.18 $\pm$ 29.22 \\
MPLight           
& 300.93 $\pm$ 3.57 
& 259.10 $\pm$ 8.90 
& 261.45 $\pm$ 1.74 
& 343.47 $\pm$ 2.89
& 282.14 $\pm$ 7.20 
& 1168.49 $\pm$ 23.99               
& 1597.24 $\pm$  18.57 \\
Advanced-CoLight 
& 253.95 $\pm$ 0.77
& 234.03 $\pm$ 0.61
& 244.51 $\pm$ 0.66
& 269.62 $\pm$ 0.51
& 308.62 $\pm$ 4.58 
& 1025.47 $\pm$ 34.33           
& -  \\
Advanced-MPLight 
& 250.04 $\pm$ 0.65
& 230.92 $\pm$ 0.36
& 227.72 $\pm$ 0.58
& 285.28 $\pm$ 0.41
& 296.76 $\pm$ 4.59 
& 1304.60 $\pm$ 18.12             
& - \\
DenseLight $*$   
& 226.97 & 215.82 & 239.58 & 248.43 & 272.27 & 803.42  & - \\
SocialLight &
217.92 $\pm$ 0.40 &
211.75 $\pm$ 0.40 &
210.46 $\pm$ 0.32 &
254.86 $\pm$ 0.23 &
288.55 $\pm$ 0.65 &
771.92 $\pm$ 5.04 &
1106.69 $\pm$ 15.39 \\ \hline
CoordLight (Ours) &
\textbf{199.24 $\pm$ 0.20} &
\textbf{198.21 $\pm$ 0.11} &
\textbf{191.05 $\pm$ 0.01} &
\textbf{248.45 $\pm$ 0.07} &
\textbf{250.87 $\pm$ 0.21} &
\textbf{748.32 $\pm$ 4.60} &
\textbf{1039.15 $\pm$ 10.05} \\ \hline

\end{tabular}%
}
\label{results}
\end{table*}

\subsection{Comparative Analysis of Average Travel Time}
First of all, we note that our experiment results in Table~\ref{results} show that CoordLight outperforms all existing baselines in reducing average travel time.
On the Jinan dataset, $\mathcal{D}_{\text{JN}}$, CoordLight is the only method with an average travel time consistently below 200 seconds. In particular, compared to the best-performing baseline method SocialLight, our approach achieves a performance improvement of 6.39$\%$, 8.57$\%$, and 9.23$\%$ over three different demands, respectively.
Furthermore, we observe that CoordLight exhibits excellent consistency and stability across three different traffic demands.
Similar conclusions can be drawn from the results of the Hangzhou dataset $\mathcal{D}_{\text{HZ}}$, where DenseLight outperforms other baseline methods across both traffic demands.
Under the low-demand dataset $\mathcal{D}_{\text{HZ}(1)}$, CoordLight achieves a performance comparable to DenseLight. However, in the higher-demand dataset $\mathcal{D}_{\text{HZ}(2)}$, CoordLight shows a 7.87$\%$ improvement over DenseLight.
Specifically, while other baseline methods struggle with the high-volume demand of $\mathcal{D}_{\text{HZ}(2)}$, our method remains stable, averaging approximately 250 seconds in performance regardless of the varying traffic intensity.

In the New York dataset $\mathcal{D}_{\text{NY}}$, the most challenging with its large scale of 196 intersections, our method surpasses both graph attention-based methods, Colight and Advanced-Colight.
We believe that this highlights how CoordLight can facilitate more efficient, targeted coordination learning by concentrating on key interactions with crucial agents within the neighborhood, compared with those cooperation methods enhanced by advanced feature extraction mechanisms (e.g., CoLight and Advanced-CoLight).
Additionally, CoordLight exhibits significant improvements over leading methods such as SocialLight and DenseLight, likely due to its advanced state representation QDSE and neighbor-aware learning algorithm NAPO that effectively enhances both local decision-making and regional coordination, thereby substantially improving traffic optimization across extensive network scales.
We performed unpaired t-tests between the results of CoordLight and SocialLight (the second best tested method) under all seven different simulation experiments, as shown in Table.~\ref{t_test}. All yielded p-values are lower than $1 \cdot 10^{-8}$; using a Bonferroni correction for these multiple tests, we find the final significance threshold $p = 1.57 \cdot 10 ^{-4}$ (for a standard, original $p = 0.01$), indicating that CoordLight (most likely) significantly outperforms SocialLight in terms of average travel time.

\begin{table*}[t]
\vspace{-0.3cm}
\centering
\caption{Unpaired T-test Results for CoordLight and SocialLight (baseline method).}
\vspace{-0.3cm}
\resizebox{15cm}{!}{%
\tiny
\begin{tabular}{cccccccc}
\hline
Traffic Dataset & $\mathcal{D}_{\text{JN(1)}}$ & $\mathcal{D}_{\text{JN(2)}}$ & $\mathcal{D}_{\text{JN(3)}}$ & $\mathcal{D}_{\text{HZ(1)}}$ & $\mathcal{D}_{\text{HZ(2)}}$ & $\mathcal{D}_{\text{NY(1)}}$ & $\mathcal{D}_{\text{NY(2)}}$ \\ \hline
P-value & 2.23e-16 & 9.99e-20 & 1.24e-21 & 8.63e-20 & 1.08e-18 & 5.45e-09 & 9.79e-09 \\ \hline
\begin{tabular}[c]{@{}c@{}}Threshold (Bonferroni Correction)\end{tabular} & \multicolumn{7}{c}{1.57e-04} \\ \hline
\end{tabular}%
}
\label{t_test}
\vspace{-0.2cm}
\end{table*}

\begin{figure}[t]
\centering
\includegraphics[width=0.95\linewidth, height=0.32\textwidth, trim=0.6cm 0.6cm 1.2cm 1.0cm, clip]{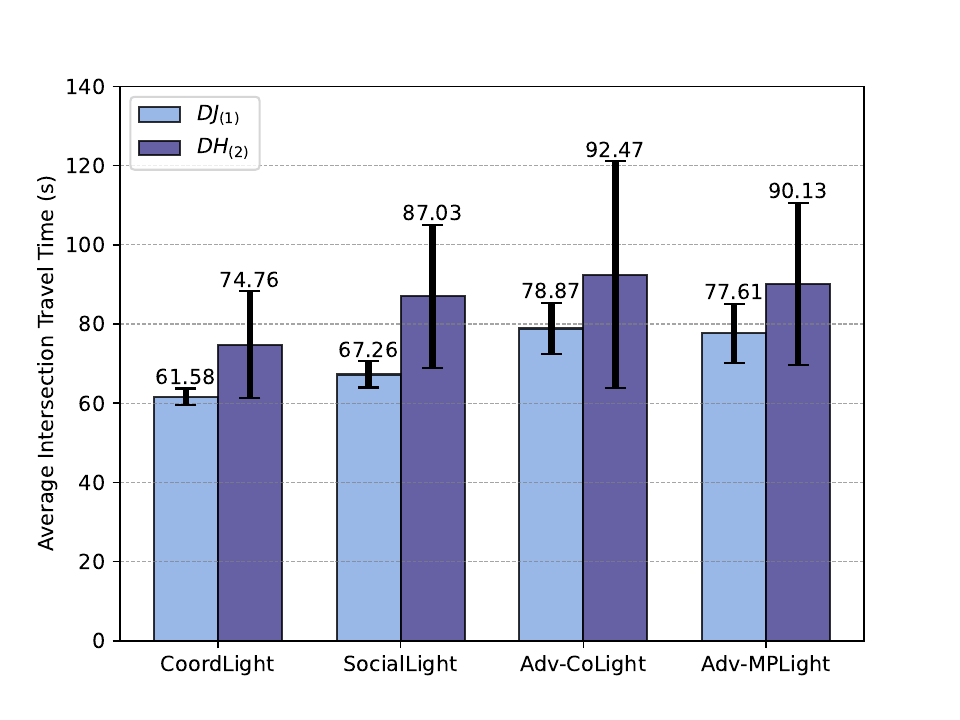}
\vspace{-0.3cm}
\caption{Comparative evaluation of the mean and standard deviation of the average travel time at intersections  (lower is better) against three MARL baselines over high-demand traffic datasets $\mathcal{D}_{\text{JN(1)}}$ and $\mathcal{D}_{\text{HZ(2)}}$.}
\label{coord_results}
\vspace{-0.6cm}
\end{figure}

\subsection{Analysis of Coordination Performance}
One of the main goals of this paper is to enhance the coordination among adjacent agents/intersections, to obtain network-wide traffic optimization through strategic traffic flow control within the overlapping neighborhood areas. 
Therefore, to investigate the coordination performance among different control methods, we study the average travel time at each intersection, as it directly reflects the efficiency of traffic flow management and the synchronization effectiveness between adjacent intersections (indicating how well agents at adjacent intersections are coordinated in timing and operation to manage traffic flow smoothly).
Specifically, lower average travel time at intersections signifies enhanced harmonization in managing interdependent traffic patterns.
This ensures that traffic flows through multiple intersections with fewer stops, ideally allowing vehicles to encounter green lights as they progress along their route.
Meanwhile, lower variance suggests better consistency and reliability in different traffic conditions.
We calculate the mean and standard deviation of travel times at each intersection using the high-demand traffic datasets $\mathcal{D}_{\text{JN(1)}}$ and $\mathcal{D}_{\text{HZ(2)}}$ during testing. The findings, illustrated in Fig.~\ref{coord_results}, reveal that CoordLight consistently achieves the lowest average travel time and variance across intersections in both datasets (yielding $61.58 \pm 2.13s$ for $\mathcal{D}_{\text{JN(1)}}$, and $74.76 \pm 13.46s$ for $\mathcal{D}_{\text{HZ(2)}}$), compared to other state-of-the-art MARL methods such as SocialLight, Advanced-CoLight and Advanced-MPLight. This implies superior coordination performance by CoordLight in optimizing network-wide traffic.

We performed additional experiments using the neighbor cooperative reward defined in SocialLight~\cite{goel2023sociallight}, as optimizing this reward requires MARL algorithms to capture key interactions and ensure accurate credit assignment for learning efficient coordination and cooperation (i.e., beyond cases involving individual rewards only). 
Detailed results can be found \href{https://www.dropbox.com/scl/fi/zus8vsyf0mm3b7ugeb1tn/Supplementary_Materials_TITS.pdf?rlkey=75ak19sltebk88l51bxrwc227&dl=0}{here}.
Our intuition about these results is that, by training with cooperative reward, we allow RL agents to account for a larger scale of regional traffic flow to further enhance their coordination and cooperation. Our results support this intuition, and highlight how CoordLight can be adapted to either use ``individual'' (regional) or more cooperative (neighborhood) rewards to learn effective coordination and cooperation at scale in a wide range of complex urban traffic networks.

\subsection{Ablation Study for State Representation}
We highlight the efficacy of our proposed state definition, QDSE, through a sequence of ablation studies comparing it against various traffic state definitions used in literature, which guide the decision-making of traffic agents. 
We train five variants of CoordLight conditioned on different lane-feature-based state definitions for 10000 episodes under identical experiment settings using the $\mathcal{D}_{\text{HZ(2)}}$ dataset, which include: 
\begin{enumerate}
    \item \textbf{Vehicle Counts (VC)}~\cite{wei2019colight, zheng2019diagnosing, zheng2019frap}: A commonly used state definition which incorporates the current phase index and vehicle counts on the incoming lanes.
    \item \textbf{General Pressure (GP)}~\cite{wei2019presslight}: Defines the state as the combination of current phase and intersection pressure.
    \item \textbf{Efficient Pressure (EP)}~\cite{wu2021efficient}: Utilizes current phase and intersection pressure within a specific effective range as the state, where the effective range is determined by maximum lane speed and phase duration.
    \item \textbf{Advanced Traffic State (ATS)}~\cite{zhang2022expression}: Defines the state as the current phase at intersections, incorporating both stopped vehicle counts and pressure, with each being calculated within the effective range.
    \item \textbf{Discrete Traffic State Encoding (DTSE)}~\cite{genders2016using}: A detailed image-like state representation that divides the intersection into multiple 6-meter grids. Each grid corresponds to a specific lane segment, with its state defined as 0 or 1 based on whether a vehicle is present.
    \item \textbf{Queue Dynamic State Encoding (QDSE, ours)}: 
    A state representation that incorporates lane features indicating current traffic conditions (stopped vehicle counts, entering vehicle counts, and leaving vehicle counts) , and estimating upcoming congestion (the position of the first moving vehicle and the vehicle counts that closely follow it).
    Similar to other state definitions, these features are obtained via the cameras at intersections.
\end{enumerate}

\begin{figure*}[htb]
\centering
\subfigure[Training curves for ablation study of state representation]{
\begin{minipage}[t]{0.48\linewidth}
\centering
\includegraphics[width=0.95\linewidth, height=0.8\linewidth]{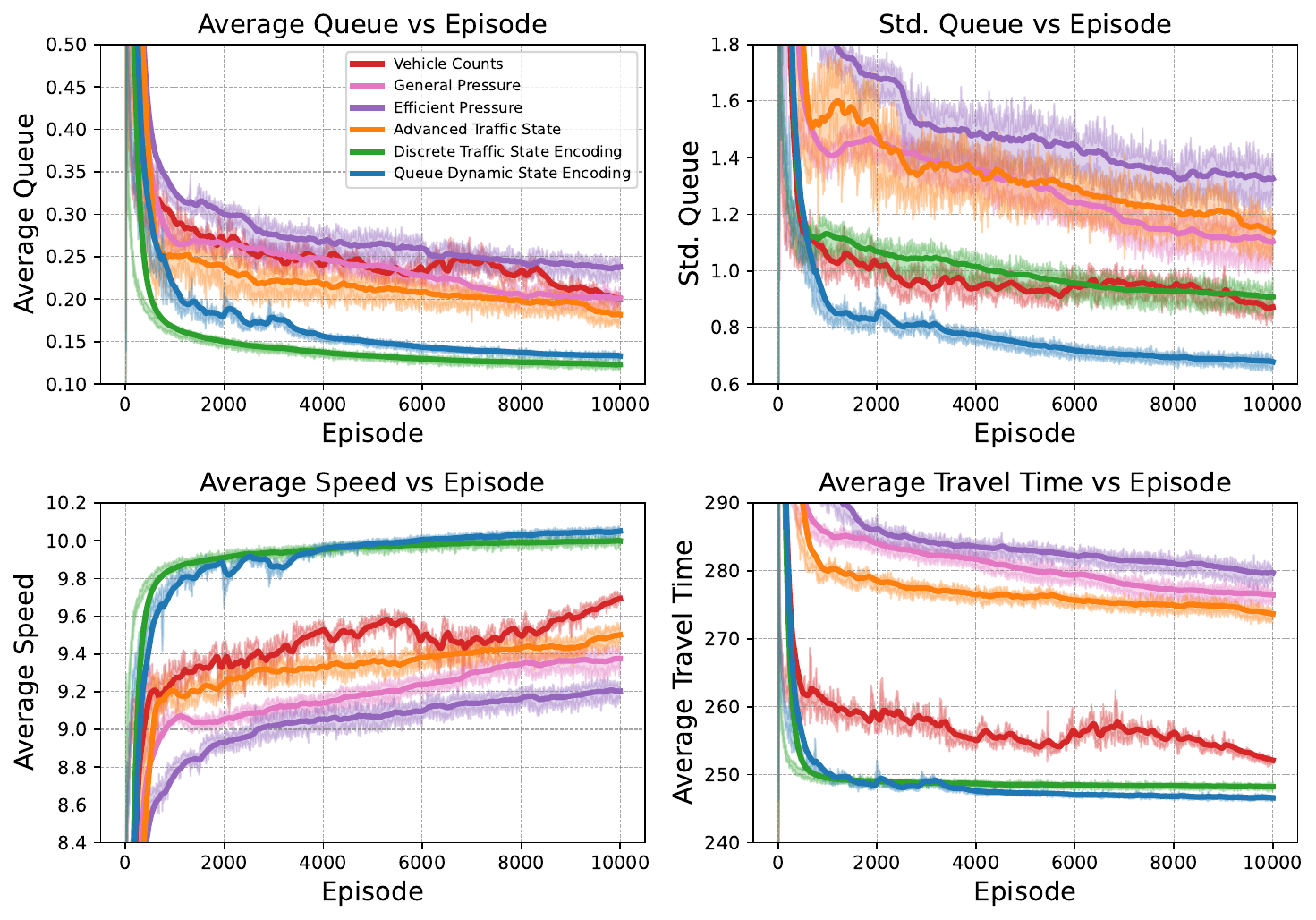}
\end{minipage}
}
\subfigure[Training curves for ablation study of the components of NAPO]{
\begin{minipage}[t]{0.48\linewidth}
\centering
\includegraphics[width=0.95\linewidth, height=0.8\linewidth]
{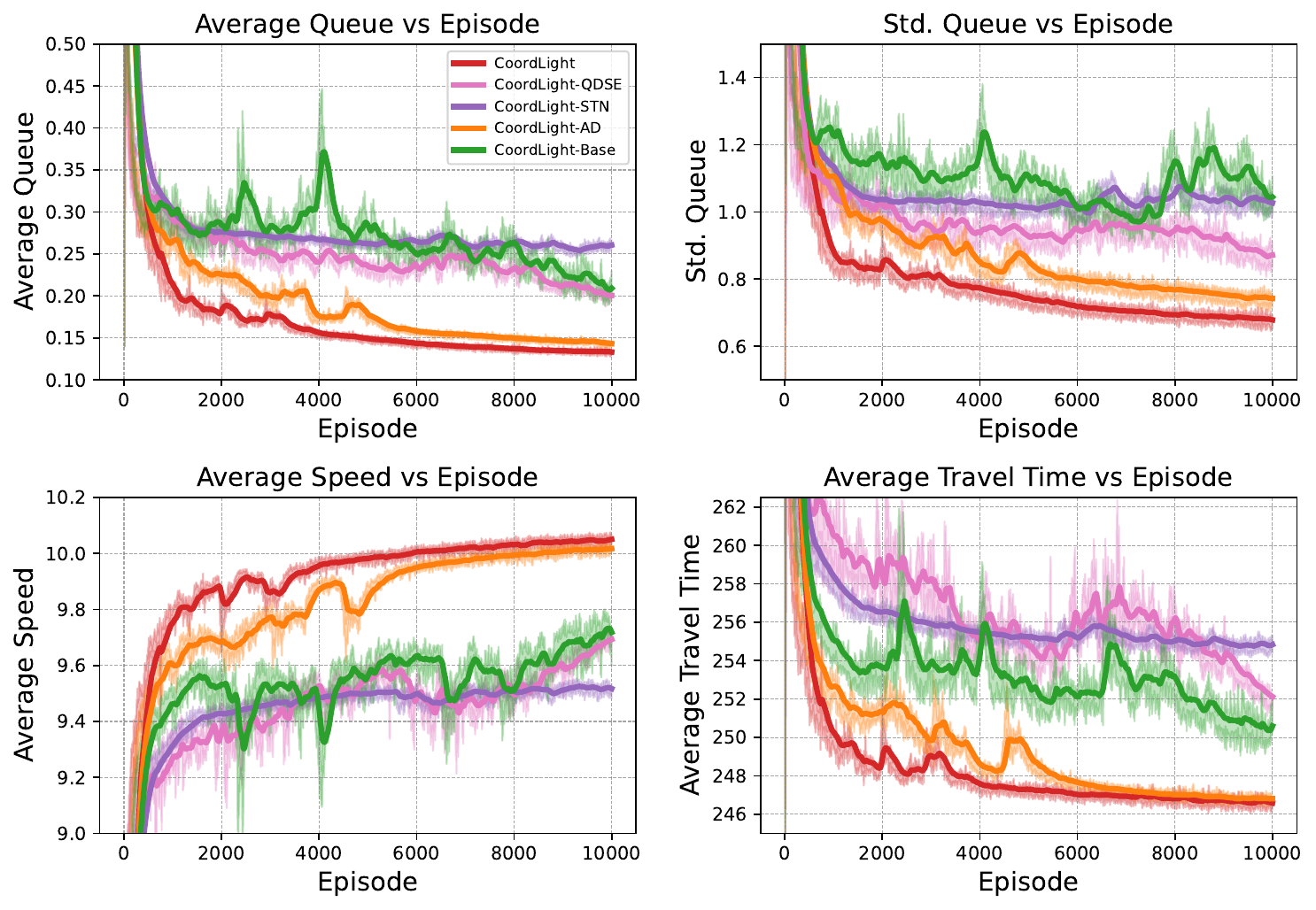}
\end{minipage}
}
\centering
\caption{Training results of the ablation variants of CoordLight in terms of average queue length ($\downarrow$), average speed ($\uparrow$), average travel time  ($\downarrow$), and standard deviation of queue length  ($\downarrow$) with respect to the $\mathcal{D}_{\text{HZ(2)}}$ dataset for 10000 episodes. Here,  $\downarrow$ indicates lower is better, while $\uparrow$ indicates higher is better.}
\vspace{-0.4cm}
\label{ablation_study}
\end{figure*}

From the training curves presented in Fig.~\ref{ablation_study}(a), we find that utilizing QDSE, as compared to other lane-feature-based state definitions, yields a notable improvement in terms of average queue length, vehicle speed, travel time and standard deviation of the queue length. Specifically, QDSE not only reduces the queue length but also significantly decreases its standard deviation. This suggests that by incorporating new lane-specific features, QDSE can effectively encapsulate the traffic dynamics, thereby mitigating myopic behaviors and enhancing the agent's predictive and responsive capabilities towards future congestion. On the other hand, the results from other state definitions based on efficient range, such as EP and ATS, indicate that only considering the vehicles that can pass the intersection within a given duration directs the agents to focus on responding to the current traffic states, neglecting the potential impact of future traffic conditions. In contrast, the features within QDSE enable agents to predict and adapt to upcoming traffic situations, rendering the control policy more proactive rather than reactive.
Comparing QDSE with the more detailed image-like state definition, we observe that using the more complex and accurate DTSE allows CoordLight to achieve smoother training curves, indicating that a more comprehensive state representation can improve training stability. Despite this, QDSE demonstrates performance on par with DTSE across all four metrics, and slightly surpasses DTSE in terms of average travel time. These results show that QDSE strikes an effective balance between complexity and accuracy by capturing the key traffic dynamics at intersections in a concise form, thereby highlighting its practicality and effectiveness toward real-world applications.

We conducted new additional experiments where we integrated QDSE with other advanced RL-based baseline methods, including IPPO~\cite{schulman2017proximal} and SocialLight~\cite{goel2023sociallight} on the most complex New York datasets $\mathcal{N}_{NY(1)}$ and $\mathcal{N}_{NY(2)}$.
By comparing the performance of these baselines with and without QDSE, we observe notable improvements over both IPPO and SocialLight when QDSE is used as the state representation. 
The results show that QDSE offers a more comprehensive depiction of intersection traffic conditions, enhancing the agent's understanding of dynamic environments and significantly reducing average travel times. We believe that these findings highlight QDSE's effectiveness in improving traffic performance for ATSC tasks and its potential to consistently enhance the performance and robustness of existing RL-based TSC algorithms. 
More detailed experimental results regarding our state representation QDSE can be found \href{https://www.dropbox.com/scl/fi/zus8vsyf0mm3b7ugeb1tn/Supplementary_Materials_TITS.pdf?rlkey=75ak19sltebk88l51bxrwc227&dl=0}{here}.

\subsection{Impacts of Sensor Noise on QDSE}
\begin{figure}[ht]
    \centering
        \includegraphics[trim=0.6cm 0.6cm 0.6cm 1.2cm, clip, width=0.45\textwidth, height=0.3\textwidth]{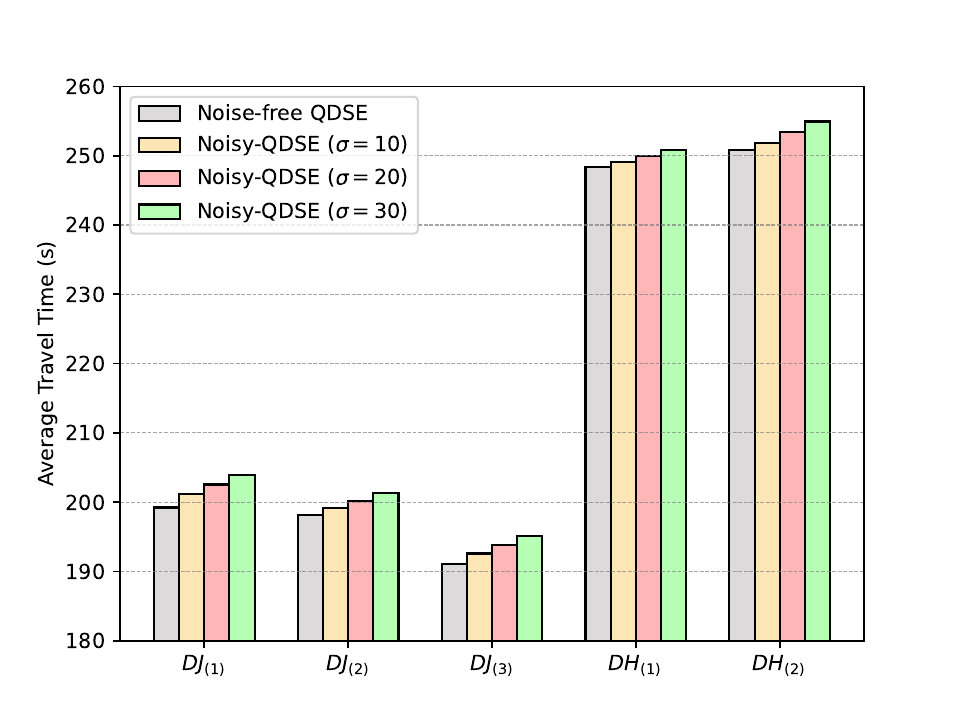}
    \vspace{-0.5cm}
    \caption{Evaluation of the proposed state representation QDSE on Jinan and Hangzhou datasets, comparing average travel time with and without sensor noise (Gaussian noise) across various traffic demands. All variants are trained with noise-free QDSE and tested with noisy-QDSE.}
    \label{noise_results}
    \vspace{-0.5cm}
\end{figure}

Given that QDSE data would typically be collected by cameras at intersections in practice, it is inevitably subject to sensor noise in real-world applications.
However, all experiments in this paper are conducted in simulators, where the state information is assumed to be collected free from sensor noise.
To evaluate the real-world applicability of QDSE, we investigate its stability and robustness under varied sensor noise conditions. We identify the measurement of the position of the leading vehicle $D_{fr}$ as the primary source of noise in QDSE. 
To simulate this, we introduce a Gaussian noise of $N(0, \sigma^2)$ to the leading vehicle's position, implying a mean of 0 and a standard deviation of $\sigma$ meter. 
Specifically, we choose $\sigma$ as $10m$, $20m$, and $30m$ to study the impacts of different levels of sensor noise on the overall control performance, respectively.
We train our models using noise-free QDSE on Jinan and Hangzhou maps with different traffic demands and then test them with noisy QDSE under identical settings across 10 episodes. 
Fig.~\ref{noise_results} demonstrates that the proposed QDSE maintains its robustness when subjected to various levels of sensor noise.
In scenarios with noise standard deviations of $10m$, $20m$, and $30m$, the performance degradation remains relatively modest, with the highest observed increase in average travel time being just over 2$\%$. Notably, the increments in degradation are gradual as the noise level increases, suggesting a consistent resilience of QDSE against escalating noise intensities. 
For instance, in the $\mathcal{D}_{\text{JN(1)}}$, the degradation rises from approximately 0.95$\%$ at $10m$ to about 2.34$\%$ at $30m$. 
This pattern suggests that, despite the adverse effects of noise, our state representation method - QDSE retains its effectiveness in environments with variable and unpredictable sensor noise, highlighting its potential for deployment in real-world scenarios.

\vspace{-0.2cm}

\subsection{Ablation Study for Proposed Components}
We systematically examine the individual contributions of various components of our CoordLight learning framework using the Hangzhou dataset $\mathcal{D}_{\text{HZ(2)}}$ with consistent experimental settings across five model variants. We train each variant for 10000 episodes, keeping hyperparameters constant for fairness. Fig.~\ref{ablation_study}.(b) displays training curves, highlighting key metrics such as mean queue length, speed, travel time, and queue length's standard deviation over training progress.
\begin{enumerate}
\item \textbf{CoordLight}: The complete independent learning framework which adopts QDSE for state representation and utilizes NAPO as its optimization algorithm.
\item \textbf{CoordLight w/o QDSE}: Substitutes the QDSE state with the Vehicle Count (VC) state, while retaining all other components unchanged.
\item \textbf{CoordLight w/o STN}: Excludes the Spatio-Temporal Network (STN) from both actor and critic network while keeping other components. As a result, it lacks the ability to capture the spatial and temporal state dependencies among adjacent agents.
\item \textbf{CoordLight w/o AD}: Removes the State-action Decoder (AD) from the critic network and retains all other components, which means it is unable to learn the state-driven action dependencies among neighboring agents.
\item \textbf{CoordLight-Base}: Utilizes a actor-critic model composed solely of fully connected layers optimized by Independent Proximal Policy Optimization (IPPO).
\end{enumerate}

Firstly, we note a significant improvement across all evaluation metrics when transitioning the state definition from a simple vehicle count to our proposed QDSE. Notably, a substantial reduction in the standard deviation of the queue length suggests that QDSE effectively amplifies the model's ability to comprehend, predict, and act on traffic dynamics, thereby alleviating potential congestion.
Secondly, after omitting the spatio-temporal network from CoordLight, the model rapidly converges to a sub-optimal solution, resulting in a remarkable decline in performance. We deduce that this occurs because the model struggles to capture key traffic dynamics and intra-neighborhood state dependencies, even with a comprehensive state representation, which impedes the learning of efficient control strategies.
Thirdly, we note that integrating action histories into value estimates appears to substantially accelerate and stabilize training in dynamic multi-agent environments.
More specifically, the state-action decoder, by assigning varied importance to neighbors' action trajectories, effectively decouples intra-neighborhood impacts and captures the dependencies among agents, thereby enhancing overall performance.
Finally, in comparison to the variant based on simple yet scalable independent learning approach (IPPO), our method, CoordLight, exhibits significant improvements across all metrics. This success is most likely attributed to its ability to learn critical state-action dependencies among neighboring agents, which significantly reduces myopic behaviors and improves the coordination awareness of the agents.

\section{Conclusion}
\label{conclusion}
In this paper, we present CoordLight, a MARL-based framework designed for adaptive traffic signal control in ever-expanding urban networks.
We first introduce an innovative state representation, Queue Dynamic State Encoding (QDSE), derived from vehicle queueing dynamics, aimed at enhancing the agents' capability to interpret, predict, and act upon local traffic dynamics.
We further propose an augmented MARL algorithm called Neighbor-aware Policy Optimization (NAPO), which enhances both decision-making and policy learning by incorporating the learning of state-action dependencies among neighboring agents via a specialized attention-based neighbor-aware actor-critic network.
These enhancements improve both the decentralized agents' local decision-making and their regional coordination, thereby boosting the overall network-wide traffic optimization. 
We conduct comprehensive experiments on three real traffic datasets composed of up to 196 intersections, and demonstrate that CoordLight consistently outperforms established state-of-the-art ATSC methods across diverse traffic networks and demand scenarios.

In response to the increasing traffic demands in metropolitan regions, advanced adaptive traffic control systems, like CoordLight, have become indispensable to ensuring efficient traffic flow management, reducing congestion, and enhancing the overall urban mobility experience.
While this study has focused on traffic optimization in homogeneous networks by selecting/sequencing fixed-duration phases, we note that our algorithm is flexible enough to be easily adapted and extended to different action spaces, such as determining phase duration or deciding whether to maintain or change the current phase. Our future work aims to extend our method to heterogeneous networks with diverse intersection layouts, lane configurations, and more realistic, asynchronous signal control settings. We believe that incorporating our QDSE and NAPO into advanced general learning frameworks can effectively enhance agent coordination and cooperation, thus leading to improved traffic performance across heterogeneous networks. Tackling these more complex and realistic traffic scenarios will be key toward achieving broader applicability in real-world environments.
Additionally, future research endeavors will focus on handling imperfect sensor data, managing traffic priorities (e.g., emergency vehicles), and navigating scenarios such as traffic accidents and road closures. These efforts aim to further bolster CoordLight's robustness, scalability, and adaptability across a wide range of varied real-life traffic conditions.

\bibliographystyle{ieeetr}
\bibliography{sample}

@misc{coma,
  doi = {10.48550/ARXIV.1705.08926},
  
  url = {https://arxiv.org/abs/1705.08926},
  
  author = {Foerster, Jakob and Farquhar, Gregory and Afouras, Triantafyllos and Nardelli, Nantas and Whiteson, Shimon},
  
  title = {Counterfactual Multi-Agent Policy Gradients},
  
  publisher = {arXiv},
  
  year = {2017},
  
  copyright = {arXiv.org perpetual, non-exclusive license}
}

@inproceedings{chen2020toward,
  title={Toward a thousand lights: Decentralized deep reinforcement learning for large-scale traffic signal control},
  author={Chen, Chacha and Wei, Hua and Xu, Nan and Zheng, Guanjie and Yang, Ming and Xiong, Yuanhao and Xu, Kai and Li, Zhenhui},
  booktitle={Proceedings of the AAAI Conference on Artificial Intelligence},
  volume={34},
  number={04},
  pages={3414--3421},
  year={2020}
}

@misc{ma2c,
  doi = {10.48550/ARXIV.1903.04527},
  
  url = {https://arxiv.org/abs/1903.04527},
  
  author = {Chu, Tianshu and Wang, Jie and Codecà, Lara and Li, Zhaojian},
  
  title = {Multi-Agent Deep Reinforcement Learning for Large-scale Traffic Signal Control},
  
  publisher = {arXiv},
  
  year = {2019},
  
  copyright = {arXiv.org perpetual, non-exclusive license}
}

@misc{attentionlight,
  doi = {10.48550/ARXIV.2201.00006},
  
  url = {https://arxiv.org/abs/2201.00006},
  
  author = {Zhang, Liang and Wu, Qiang and Deng, Jianming},
  
  title = {AttentionLight: Rethinking queue length and attention mechanism for traffic signal control},
  
  publisher = {arXiv},
  
  year = {2022},
  
  copyright = {arXiv.org perpetual, non-exclusive license}
}

@article{wu2021efficient,
  title={Efficient pressure: Improving efficiency for signalized intersections},
  author={Wu, Qiang and Zhang, Liang and Shen, Jun and L{\"u}, Linyuan and Du, Bo and Wu, Jianqing},
  journal={arXiv preprint arXiv:2112.02336},
  year={2021}
}

@inproceedings{tang2019cityflow,
  title={Cityflow: A city-scale benchmark for multi-target multi-camera vehicle tracking and re-identification},
  author={Tang, Zheng and Naphade, Milind and Liu, Ming-Yu and Yang, Xiaodong and Birchfield, Stan and Wang, Shuo and Kumar, Ratnesh and Anastasiu, David and Hwang, Jenq-Neng},
  booktitle={Proceedings of the IEEE/CVF Conference on Computer Vision and Pattern Recognition},
  pages={8797--8806},
  year={2019}
}

@inproceedings{wei2019colight,
      title={Colight: Learning network-level cooperation for traffic signal control},
      author={Wei, Hua and Xu, Nan and Zhang, Huichu and Zheng, Guanjie and Zang, Xinshi and Chen, Chacha and Zhang, Weinan and Zhu, Yanmin and Xu, Kai and Li, Zhenhui},
      booktitle={Proceedings of the 28th ACM International Conference on Information and Knowledge Management},
      pages={1913--1922},
      year={2019}
    }

@inproceedings{zheng2019frap,
      title={Learning phase competition for traffic signal control},
      author={Zheng, Guanjie and Xiong, Yuanhao and Zang, Xinshi and Feng, Jie and Wei, Hua and Zhang, Huichu and Li, Yong and Xu, Kai and Li, Zhenhui},
      booktitle={Proceedings of the 28th ACM International Conference on Information and Knowledge Management},
      pages={1963--1972},
      year={2019}
    }

@techreport{koonce2008traffic,
  title={Traffic signal timing manual.},
  author={Koonce, Peter and Rodegerdts, Lee},
  year={2008},
  institution={United States. Federal Highway Administration}
}

@book{roess2004traffic,
  title={Traffic engineering},
  author={Roess, Roger P and Prassas, Elena S and McShane, William R},
  year={2004},
  publisher={Pearson/Prentice Hall}
}

@article{little1981maxband,
  title={MAXBAND: A versatile program for setting signals on arteries and triangular networks},
  author={Little, John DC and Kelson, Mark D and Gartner, Nathan H},
  year={1981},
  publisher={Alfred P. Sloan School of Management, Massachusetts Institute of Technology}
}

@incollection{varaiya2013max,
  title={The max-pressure controller for arbitrary networks of signalized intersections},
  author={Varaiya, Pravin},
  booktitle={Advances in dynamic network modeling in complex transportation systems},
  pages={27--66},
  year={2013},
  publisher={Springer}
}

@article{lowrie1990scats,
  title={Scats, sydney co-ordinated adaptive traffic system: A traffic responsive method of controlling urban traffic},
  author={Lowrie, PR},
  year={1990}
}

@article{li2016traffic,
  title={Traffic signal timing via deep reinforcement learning},
  author={Li, Li and Lv, Yisheng and Wang, Fei-Yue},
  journal={IEEE/CAA Journal of Automatica Sinica},
  volume={3},
  number={3},
  pages={247--254},
  year={2016},
  publisher={IEEE}
}

@inproceedings{garg2018deep,
  title={Deep reinforcement learning for autonomous traffic light control},
  author={Garg, Deepeka and Chli, Maria and Vogiatzis, George},
  booktitle={2018 3rd IEEE international conference on intelligent transportation engineering (ICITE)},
  pages={214--218},
  year={2018},
  organization={IEEE}
}

@article{liang2019deep,
  title={A deep reinforcement learning network for traffic light cycle control},
  author={Liang, Xiaoyuan and Du, Xunsheng and Wang, Guiling and Han, Zhu},
  journal={IEEE Transactions on Vehicular Technology},
  volume={68},
  number={2},
  pages={1243--1253},
  year={2019},
  publisher={IEEE}
}

@inproceedings{prashanth2011reinforcement,
  title={Reinforcement learning with average cost for adaptive control of traffic lights at intersections},
  author={Prashanth, LA and Bhatnagar, Shalabh},
  booktitle={2011 14th International IEEE Conference on Intelligent Transportation Systems (ITSC)},
  pages={1640--1645},
  year={2011},
  organization={IEEE}
}

@article{balaji2010urban,
  title={Urban traffic signal control using reinforcement learning agents},
  author={Balaji, PG and German, X and Srinivasan, Dipti},
  journal={IET Intelligent Transport Systems},
  volume={4},
  number={3},
  pages={177--188},
  year={2010},
  publisher={IET}
}

@inproceedings{xiong2019learning,
  title={Learning traffic signal control from demonstrations},
  author={Xiong, Yuanhao and Zheng, Guanjie and Xu, Kai and Li, Zhenhui},
  booktitle={Proceedings of the 28th ACM International Conference on Information and Knowledge Management},
  pages={2289--2292},
  year={2019}
}

@inproceedings{mnih2016asynchronous,
  title={Asynchronous methods for deep reinforcement learning},
  author={Mnih, Volodymyr and Badia, Adria Puigdomenech and Mirza, Mehdi and Graves, Alex and Lillicrap, Timothy and Harley, Tim and Silver, David and Kavukcuoglu, Koray},
  booktitle={ICML},
  pages={1928--1937},
  year={2016},
  organization={PMLR}
}

@article{mousavi2017traffic,
  title={Traffic light control using deep policy-gradient and value-function-based reinforcement learning},
  author={Mousavi, Seyed Sajad and Schukat, Michael and Howley, Enda},
  journal={IET Intelligent Transport Systems},
  volume={11},
  number={7},
  pages={417--423},
  year={2017},
  publisher={IET}
}

@inproceedings{xie2020iedqn,
  title={Iedqn: Information exchange dqn with a centralized coordinator for traffic signal control},
  author={Xie, Donghan and Wang, Zhi and Chen, Chunlin and Dong, Daoyi},
  booktitle={2020 International Joint Conference on Neural Networks (IJCNN)},
  pages={1--8},
  year={2020},
  organization={IEEE}
}

@inproceedings{wei2019presslight,
  title={Presslight: Learning max pressure control to coordinate traffic signals in arterial network},
  author={Wei, Hua and Chen, Chacha and Zheng, Guanjie and Wu, Kan and Gayah, Vikash and Xu, Kai and Li, Zhenhui},
  booktitle={Proceedings of the 25th ACM SIGKDD International Conference on Knowledge Discovery \& Data Mining},
  pages={1290--1298},
  year={2019}
}

@article{wang2020stmarl,
  title={STMARL: A spatio-temporal multi-agent reinforcement learning approach for cooperative traffic light control},
  author={Wang, Yanan and Xu, Tong and Niu, Xin and Tan, Chang and Chen, Enhong and Xiong, Hui},
  journal={IEEE Transactions on Mobile Computing},
  volume={21},
  number={6},
  pages={2228--2242},
  year={2020},
  publisher={IEEE}
}

@article{chu2019multi,
  title={Multi-agent deep reinforcement learning for large-scale traffic signal control},
  author={Chu, Tianshu and Wang, Jie and Codec{\`a}, Lara and Li, Zhaojian},
  journal={IEEE Transactions on Intelligent Transportation Systems},
  volume={21},
  number={3},
  pages={1086--1095},
  year={2019},
  publisher={IEEE}
}

@article{ma2021deep,
  title={A deep reinforcement learning approach to traffic signal control with temporal traffic pattern mining},
  author={Ma, Dongfang and Zhou, Bin and Song, Xiang and Dai, Hanwen},
  journal={IEEE Transactions on Intelligent Transportation Systems},
  volume={23},
  number={8},
  pages={11789--11800},
  year={2021},
  publisher={IEEE}
}

@article{zhou2024cooperative,
  title={Cooperative Traffic Signal Control Using a Distributed Agent-Based Deep Reinforcement Learning With Incentive Communication},
  author={Zhou, Bin and Zhou, Qishen and Hu, Simon and Ma, Dongfang and Jin, Sheng and Lee, Der-Horng},
  journal={IEEE Transactions on Intelligent Transportation Systems},
  year={2024},
  publisher={IEEE}
}

@article{song2024cooperative,
  title={Cooperative traffic signal control through a counterfactual multi-agent deep actor critic approach},
  author={Song, Xiang Ben and Zhou, Bin and Ma, Dongfang},
  journal={Transportation Research Part C: Emerging Technologies},
  volume={160},
  pages={104528},
  year={2024},
  publisher={Elsevier}
}

@inproceedings{wei2018intellilight,
  title={Intellilight: A reinforcement learning approach for intelligent traffic light control},
  author={Wei, Hua and Zheng, Guanjie and Yao, Huaxiu and Li, Zhenhui},
  booktitle={Proceedings of the 24th ACM SIGKDD International Conference on Knowledge Discovery \& Data Mining},
  pages={2496--2505},
  year={2018}
}

@article{genders2016using,
  title={Using a deep reinforcement learning agent for traffic signal control},
  author={Genders, Wade and Razavi, Saiedeh},
  journal={arXiv preprint arXiv:1611.01142},
  year={2016}
}

@inproceedings{gupta2017cooperative,
  title={Cooperative multi-agent control using deep reinforcement learning},
  author={Gupta, Jayesh K and Egorov, Maxim and Kochenderfer, Mykel},
  booktitle={AAMAS},
  pages={66--83},
  year={2017},
  organization={Springer}
}

@article{liu2021learning,
  title={Learning scalable multi-agent coordination by spatial differentiation for traffic signal control},
  author={Liu, Junjia and Zhang, Huimin and Fu, Zhuang and Wang, Yao},
  journal={Engineering Applications of Artificial Intelligence},
  volume={100},
  pages={104165},
  year={2021},
  publisher={Elsevier}
}

@article{de2020independent,
  title={Is independent learning all you need in the starcraft multi-agent challenge?},
  author={de Witt, Christian Schroeder and Gupta, Tarun and Makoviichuk, Denys and Makoviychuk, Viktor and Torr, Philip HS and Sun, Mingfei and Whiteson, Shimon},
  journal={arXiv preprint arXiv:2011.09533},
  year={2020}
}

@article{schulman2017proximal,
  title={Proximal policy optimization algorithms},
  author={Schulman, John and Wolski, Filip and Dhariwal, Prafulla and Radford, Alec and Klimov, Oleg},
  journal={arXiv preprint arXiv:1707.06347},
  year={2017}
}

@article{damani2021primal,
  title={PRIMAL $ \_2 $: Pathfinding via reinforcement and imitation multi-agent learning-lifelong},
  author={Damani, Mehul and Luo, Zhiyao and Wenzel, Emerson and Sartoretti, Guillaume},
  journal={IEEE Robotics and Automation Letters},
  volume={6},
  number={2},
  pages={2666--2673},
  year={2021},
  publisher={IEEE}
}

@misc{mappo,
  doi = {10.48550/ARXIV.2103.01955},
  
  url = {https://arxiv.org/abs/2103.01955},
  
  author = {Yu, Chao and Velu, Akash and Vinitsky, Eugene and Wang, Yu and Bayen, Alexandre and Wu, Yi},
  
  title = {The Surprising Effectiveness of PPO in Cooperative, Multi-Agent Games},
  
  publisher = {arXiv},
  
  year = {2021},
  
  copyright = {arXiv.org perpetual, non-exclusive license}
}

@inproceedings{rashid2018qmix,
  title={Qmix: Monotonic value function factorisation for deep multi-agent reinforcement learning},
  author={Rashid, Tabish and Samvelyan, Mikayel and Schroeder, Christian and Farquhar, Gregory and Foerster, Jakob and Whiteson, Shimon},
  booktitle={ICML},
  pages={4295--4304},
  year={2018}
}

@inproceedings{iqbal2019actor,
  title={Actor-attention-critic for multi-agent reinforcement learning},
  author={Iqbal, Shariq and Sha, Fei},
  booktitle={ICML},
  pages={2961--2970},
  year={2019},
  organization={PMLR}
}

@article{gronauer2022multi,
  title={Multi-agent deep reinforcement learning: a survey},
  author={Gronauer, Sven and Diepold, Klaus},
  journal={Artificial Intelligence Review},
  volume={55},
  number={2},
  pages={895--943},
  year={2022},
  publisher={Springer}
}

@article{zheng2019diagnosing,
  title={Diagnosing reinforcement learning for traffic signal control},
  author={Zheng, Guanjie and Zang, Xinshi and Xu, Nan and Wei, Hua and Yu, Zhengyao and Gayah, Vikash and Xu, Kai and Li, Zhenhui},
  journal={arXiv preprint arXiv:1905.04716},
  year={2019}
}

@article{treiber2000congested,
  title={Congested traffic states in empirical observations and microscopic simulations},
  author={Treiber, Martin and Hennecke, Ansgar and Helbing, Dirk},
  journal={Physical review E},
  volume={62},
  number={2},
  pages={1805},
  year={2000},
  publisher={APS}
}

@article{chung2014empirical,
  title={Empirical evaluation of gated recurrent neural networks on sequence modeling},
  author={Chung, Junyoung and Gulcehre, Caglar and Cho, KyungHyun and Bengio, Yoshua},
  journal={arXiv preprint arXiv:1412.3555},
  year={2014}
}

@inproceedings{zhang2022expression,
  title={Expression might be enough: Representing pressure and demand for reinforcement learning based traffic signal control},
  author={Zhang, Liang and Wu, Qiang and Shen, Jun and L{\"u}, Linyuan and Du, Bo and Wu, Jianqing},
  booktitle={ICML},
  pages={26645--26654},
  year={2022},
  organization={PMLR}
}

@article{lin2023denselight,
  title={DenseLight: Efficient Control for Large-scale Traffic Signals with Dense Feedback},
  author={Lin, Junfan and Zhu, Yuying and Liu, Lingbo and Liu, Yang and Li, Guanbin and Lin, Liang},
  journal={arXiv preprint arXiv:2306.07553},
  year={2023}
}

@article{goel2023sociallight,
  title={SocialLight: Distributed Cooperation Learning towards Network-Wide Traffic Signal Control},
  author={Goel, Harsh and Zhang, Yifeng and Damani, Mehul and Sartoretti, Guillaume},
  journal={arXiv preprint arXiv:2305.16145},
  year={2023}
}

@article{vaswani2017attention,
  title={Attention is all you need},
  author={Vaswani, Ashish and Shazeer, Noam and Parmar, Niki and Uszkoreit, Jakob and Jones, Llion and Gomez, Aidan N and Kaiser, {\L}ukasz and Polosukhin, Illia},
  journal={Advances in neural information processing systems},
  volume={30},
  year={2017}
}

@article{chu2020multi,
  title={Multi-agent reinforcement learning for networked system control},
  author={Chu, Tianshu and Chinchali, Sandeep and Katti, Sachin},
  journal={arXiv preprint arXiv:2004.01339},
  year={2020}
}

@article{zhang2022neighborhood,
  title={Neighborhood cooperative multiagent reinforcement learning for adaptive traffic signal control in epidemic regions},
  author={Zhang, Chengwei and Tian, Yu and Zhang, Zhibin and Xue, Wanli and Xie, Xiaofei and Yang, Tianpei and Ge, Xin and Chen, Rong},
  journal={IEEE T-ITS},
  volume={23},
  number={12},
  pages={25157--25168},
  year={2022},
  publisher={IEEE}
}

@article{kingma2014adam,
  title={Adam: A method for stochastic optimization},
  author={Kingma, Diederik P and Ba, Jimmy},
  journal={arXiv preprint arXiv:1412.6980},
  year={2014}
}

@article{schulman2015high,
  title={High-dimensional continuous control using generalized advantage estimation},
  author={Schulman, John and Moritz, Philipp and Levine, Sergey and Jordan, Michael and Abbeel, Pieter},
  journal={arXiv preprint arXiv:1506.02438},
  year={2015}
}

@article{aslani2017adaptive,
  title={Adaptive traffic signal control with actor-critic methods in a real-world traffic network with different traffic disruption events},
  author={Aslani, Mohammad and Mesgari, Mohammad Saadi and Wiering, Marco},
  journal={Transportation Research Part C: Emerging Technologies},
  volume={85},
  pages={732--752},
  year={2017},
  publisher={Elsevier}
}

@article{aslani2018traffic,
  title={Traffic signal optimization through discrete and continuous reinforcement learning with robustness analysis in downtown Tehran},
  author={Aslani, Mohammad and Seipel, Stefan and Mesgari, Mohammad Saadi and Wiering, Marco},
  journal={Advanced Engineering Informatics},
  volume={38},
  pages={639--655},
  year={2018},
  publisher={Elsevier}
}

@inproceedings{abdoos2011traffic,
  title={Traffic light control in non-stationary environments based on multi agent Q-learning},
  author={Abdoos, Monireh and Mozayani, Nasser and Bazzan, Ana LC},
  booktitle={2011 14th International IEEE conference on intelligent transportation systems (ITSC)},
  pages={1580--1585},
  year={2011},
  organization={IEEE}
}

@article{casas2017deep,
  title={Deep deterministic policy gradient for urban traffic light control},
  author={Casas, Noe},
  journal={arXiv preprint arXiv:1703.09035},
  year={2017}
}

@article{brys2014distributed,
  title={Distributed learning and multi-objectivity in traffic light control},
  author={Brys, Tim and Pham, Tong T and Taylor, Matthew E},
  journal={Connection Science},
  volume={26},
  number={1},
  pages={65--83},
  year={2014},
  publisher={Taylor \& Francis}
}

@article{mannion2016experimental,
  title={An experimental review of reinforcement learning algorithms for adaptive traffic signal control},
  author={Mannion, Patrick and Duggan, Jim and Howley, Enda},
  journal={Autonomic road transport support systems},
  pages={47--66},
  year={2016},
  publisher={Springer}
}

@article{van2016coordinated,
  title={Coordinated deep reinforcement learners for traffic light control},
  author={Van der Pol, Elise and Oliehoek, Frans A},
  journal={Proceedings of learning, inference and control of multi-agent systems (at NIPS 2016)},
  volume={8},
  pages={21--38},
  year={2016}
}

@article{lin2023temporal,
  title={Temporal Difference-Aware Graph Convolutional Reinforcement Learning for Multi-Intersection Traffic Signal Control},
  author={Lin, Wei-Yu and Song, Yun-Zhu and Ruan, Bo-Kai and Shuai, Hong-Han and Shen, Chih-Ya and Wang, Li-Chun and Li, Yung-Hui},
  journal={IEEE Transactions on Intelligent Transportation Systems},
  year={2023},
  publisher={IEEE}
}

@article{zhu2023metavim,
  title={Metavim: Meta variationally intrinsic motivated reinforcement learning for decentralized traffic signal control},
  author={Zhu, Liwen and Peng, Peixi and Lu, Zongqing and Tian, Yonghong},
  journal={IEEE Transactions on Knowledge and Data Engineering},
  volume={35},
  number={11},
  pages={11570--11584},
  year={2023},
  publisher={IEEE}
}

@inproceedings{chen2017decentralized,
  title={Decentralized non-communicating multiagent collision avoidance with deep reinforcement learning},
  author={Chen, Yu Fan and Liu, Miao and Everett, Michael and How, Jonathan P},
  booktitle={2017 IEEE international conference on robotics and automation (ICRA)},
  pages={285--292},
  year={2017},
  organization={IEEE}
}

\end{document}